# Explainable AI for Predicting and Understanding Mathematics Achievement:
## A Cross-National Analysis of PISA 2018

Liu Liu[1*], Dai Rui[2*]

[1] College of Education, University of Washington - Seattle
[2] Arizona State University

**\*** Corresponding Author:
Liu Liu, Email: liuuil@uga.edu
Rui Dai, Email: rdai5@asu.edu

**Date:** August 2025

## Abstract

Understanding the multifaceted factors that influence students' mathematics performance is important for designing effective educational policies and targeted interventions. This study aims to apply explainable artificial intelligence (XAI) techniques to PISA 2018 data to better predict math achievement and uncover key predictors across diverse national contexts. We analyzed data for 67,329 students from ten countries in PISA 2018, spanning Asia, Europe, North America, and Latin America that vary widely in cultural and economic contexts. We developed four predictive models, including Multiple Linear Regression (MLR), Random Forest (RF), CATBoost, and Artificial Neural Network (ANN), to estimate students' maths scores using a broad set of student-, family-, and school-level variables. Models were trained on 70% of the data (with 5-fold cross-validation for hyperparameter tuning) and tested on the remaining 30%, stratified by country. We evaluated performance using the coefficient of determination ($R^2$) and Mean Absolute Error (MAE). To ensure interpretability, we employed explainable AI techniques: we examined feature importance from the ML models and used SHAP values and decision tree visualizations to understand how each predictor influenced the predicted scores. Non-linear ML models, particularly ANN and RF, significantly outperformed MLR in predictive accuracy, with RF offering a superior balance between accuracy and generalizability. Feature importance analyses revealed that socio-economic status, learning time, and engagement-related factors such as teacher motivation and students' attitudes towards mathematics were among the most influential predictors, although their prominence varied across countries. Visual diagnostics (e.g., hexagonal bin scatterplots of predicted vs. actual scores) confirmed that the ML models captured the overall performance patterns well, with RF and CatBoost showing predictions tightly aligned with actual scores. These findings underscore the non-linear and context-dependent nature of academic achievement and highlight the value of explainable AI approaches in educational research. By uncovering cross-national patterns and context-specific drivers of performance, this study offers methodological and practical contributions. It provides a data-informed foundation to support equity-oriented educational reforms and the development of personalized learning strategies.

*Keywords*: Explainable AI; PISA 2018; Mathematics Achievement; Cross-national Comparison; Machine Learning; Large-scale Assessment

**Explainable AI for Predicting and Understanding Mathematics Achievement:**

**A Cross-National Analysis of PISA 2018**

**Introduction**

Mathematics proficiency is widely recognized as a fundamental component of educational success and economic competitiveness (Ball, 2003; Loewenberg, 2003). Strong math skills correlate with overall academic success and cultivate critical thinking, problem-solving, and adaptability in technology-rich environments (Chen et al., 2018; Edens & Potter, 2013). In an increasingly knowledge-driven economy, nations view improvements in mathematics education as vital for workforce development and innovation capacity. As a result, understanding the determinants of student performance in mathematics is of paramount importance for educators and policymakers.

Extensive research has examined how various factors, including cognitive abilities, socio-economic status (SES), family background, and school environments shape mathematics achievement (e.g., Blums et al., 2017; Lee & Chen, 2019; Lü et al., 2023; Mazurek, 2019; Wang et al., 2014, 2015). These studies consistently find that SES (often measured via parental education, occupation, and home resources) is a strong predictor of math scores: students from higher-SES families tend to score significantly higher in math. Parental education and home educational resources likewise contribute to performance (e.g., more educated parents and book-rich home environments are linked to better outcomes). Student attitudes and behaviors also matter; characteristics such as self-efficacy, interest in mathematics, and perseverance show positive associations with math achievement. For example, analyses of international assessment data have found that students who are more confident and interested in math perform better, independent of their background (Lee & Stankov, 2018; Zhang & Wang, 2020). Similarly, school context plays a role: supportive school climates, adequate resources, and positive teacher–student relationships have been linked to higher achievement. Even in high-performing systems, differences in classroom climate or teacher support can impact student outcomes (Schleicher, 2019; Üstün et al., 2022). Taken together, prior research highlights that students' mathematics achievement is complex combination of individual dispositions, family background, and school factors. Many of these influences are interrelated and may interact in non-linear ways. This poses a vital analytical challenge: traditional statistical methods like ordinary least squares regression assume additive, linear effects and may oversimplify these relationships. Because linear models have difficulty capturing high-dimensional interactions, they might fail to detect how multiple factors jointly influence achievement. This limitation has paved the way for more flexible artificial intelligence (AI) and machine learning methods in educational research (e.g., Collier et al., 2022).

The Programme for International Student Assessment (PISA), conducted triennially by the Organization for Economic Co-operation and Development (OECD), is a prime example of a large-scale assessment that collects multifaceted data on student achievement and background (Schleicher, 2019). The PISA 2018 cycle tested over 600,000 15-year-olds across 79 countries in reading, mathematics, and science. Prior studies using PISA data have predominantly employed linear or multilevel models to identify predictors of achievement such as family SES, school resources, and student attitudes (e.g., Pitsia et al., 2017; Sun et al., 2012). While these approaches have yield valuable insights (for instance, confirming that higher-SES students tend to score better), they may not fully capture the joint influence of multiple variables or reveal subtle, context-dependent interactions (e.g., Boman, 2023; Gabriel et al., 2018). The sheer scale and complexity of PISA data – with hundreds of variables and nested data structures – can outstrip the capabilities of traditional linear models, especially in cross-national analyses.

Machine learning (ML) techniques have increasingly gained traction for analyzing educational data, including international assessments like PISA. Unlike purely linear models, ML methods such as decision trees, random forests, gradient boosting machines, and neural networks can automatically detect complex interactions and non-linear patterns in the data. For example, decision tree algorithms can split students into performance groups based on combinations of attributes, potentially uncovering rules like "low parental SES and low study time yields low performance" that a linear model would miss. Prior applications of ML to PISA have shown promising results. Gamazo and Martínez-Abad (2020) used a C4.5 decision tree on PISA 2018 data and found that school-average SES was the top splitter, indicating that socio-economic context differentiated high- vs. low-performing schools. Huang et al. (2024) applied an XGBoost (extreme gradient boosting) model to PISA mathematics data and achieved very high predictive accuracy, outperforming linear models; their use of explainable boosting methods revealed mathematics self-efficacy as the single most influential predictor overall. Similarly, Hong et al. (2022) demonstrated that random forest models could incorporate both individual attitudes and school demographic factors, with each type emerging among top predictors of PISA science performance. These examples illustrate that non-linear ML methods often yield better predictive performance than traditional regressions in educational contexts.

Importantly, ML has enabled comparative analyses across countries that were previously difficult. Because ML models can handle large numbers of variables and automatically model interactions, researchers have begun using them to explore cross-national patterns in PISA. For instance, Bayirli et al. (2023) applied ML classifiers (random forests and support vector machines) to PISA 2018 data for multiple Asia-Pacific countries, finding that certain predictors (parental education, home resources, study time, age) consistently influenced achievement, though the relative importance of these predictors varied by country. Such findings highlight ML's capacity to reveal both universal predictors and context-specific patterns. In Bayirli et al.'s study, parental education was critical everywhere, but East Asian countries showed an especially strong emphasis on study diligence and early educational preparation, whereas other countries exhibited different profiles. In summary, the use of ML in large-scale assessment research has progressed from exploratory single-country studies to more sophisticated cross-national analyses. Ensemble tree methods and neural networks have proven particularly effective in modeling PISA data, often achieving higher accuracy than conventional models by capturing non-linear interactions.

A common criticism of complex ML models, however, is their "black box" nature, which limits transparency in how predictions are made (Hassija et al., 2024). In education, stakeholders require interpretable insights to inform action; an accurate model is of limited use if we cannot explain its results to teachers and policymakers (Albreiki et al., 2021). To address this, the field has seen a growing emphasis on explainable AI (XAI) techniques that improve interpretability without sacrificing much predictive power. One widely used approach is SHAP (Shapley Additive Explanations), which assigns each feature an importance value for individual predictions, showing how that feature increases or decreases a student's predicted score (Lundberg & Lee, 2017). For instance, using SHAP we might explain that, for a particular student, coming from a low-SES background reduced their predicted math score by 20 points, whereas spending more time learning math added 15 points. Recent educational ML studies have incorporated such tools: Huang et al. (2024) used SHAP to interpret their PISA model and identified key influences like self-efficacy, quantifying each factor's contribution. Decision trees themselves offer inherent interpretability; Gamazo and Martínez-Abad (2020) visualized decision rules (e.g., splits on SES) from their tree to make the model's reasoning accessible to educators. Other methods include feature importance plots (which rank variables by predictive power) and rule extraction from complex models. While some models (e.g., simple decision trees or linear models) are inherently interpretable, more complex ones (neural nets, ensembles) require post-hoc explanation techniques. Balancing the trade-off between accuracy and interpretability remains a central concern in ML for education. Fortunately, tools like SHAP, partial dependence plots, and surrogate decision trees can bridge this gap, allowing us to harness complex models while still deriving actionable insights (Hassija et al., 2024; Immekus et al., 2022).

 Despite the advances, several important gaps persist in the literature. First, most ML studies on educational data have been region-specific or single-country, which limits generalizability. Findings from one country (or a set of similar countries) may not hold elsewhere due to cultural and systemic differences. For example, some have examined PISA data within one country or region (Lee, 2022 – China; Ertem, 2021 – Turkey; Bayirli et al., 2023 – Asia-Pacific). While these provide valuable insights, we still lack truly global perspectives testing whether certain predictors (like motivation) are universally important or context-dependent. Cross-continental comparisons using ML are needed to evaluate the generality of predictive factors. Second, many studies have relied on a single type of model, often a random forest, without benchmarking against other approaches like boosting or neural networks. It remains underexplored which ML algorithm is "best" for large-scale educational predictions, as direct comparisons are few. Systematic head-to-head evaluations of multiple ML algorithms on the same data are still relatively rare in education (Wang et al., 2023 noted this gap). This includes a paucity of research incorporating deep learning models like neural networks in PISA analyses – possibly due to their complexity – leaving open the question of whether more flexible neural nets can capture patterns in education data better than structured models. Third, model interpretability is frequently treated as secondary to predictive performance. While some ML studies list the top predictors from their models, they rarely use advanced explanation techniques like SHAP to delve deeper into how features interact or contribute. Conversely, studies that prioritize interpretability often stick to simpler models, possibly sacrificing accuracy. There is a need for approaches that strike a balance that delivers strong predictive performance while remaining interpretable and relevant to practitioners. In summary, the existing literature calls for

(1) more cross-national ML studies to test the generality of findings, (2) direct comparisons of multiple ML algorithms on the same educational dataset, and (3) greater integration of explainable AI techniques so that results are transparent and actionable. These gaps inform the motivation and design of the present study.

In this study, we address these gaps by applying state-of-the-art ML models to the publicly available PISA 2018 mathematics dataset from ten countries that represent a wide range of educational contexts. We incorporate explainability techniques (such as feature importance analysis, SHAP values, and decision tree visualizations) to reveal how different factors contribute to student performance in each country. By using a consistent set of predictors and models across all countries, we can identify both generalizable patterns and context-specific influences in math achievement. Throughout the analysis, we emphasize interpretability to ensure the results are understandable and useful for the education community (educators, policymakers, and researchers). The goals of this work are both methodological and practical: we demonstrate how explainable AI can improve the prediction of student outcomes while maintaining transparency, and we derive insights about the drivers of math achievement that can inform educational policy and practice. In doing so, we present an approach for using large-scale educational data that balances predictive accuracy with explanation, providing a model for future research in educational data science.

## Method

### Data and Sample

This study uses data from the PISA 2018 assessment, focusing on student performance in mathematics. To manage scope while ensuring diversity, we analyzed a subset comprising 67,329 students from ten countries: Argentina, Chile, Chinese Taipei (Taiwan), Finland, Hungary, Italy, Japan, South Korea, the Philippines, and the United States. These countries were selected to represent a broad range of educational systems, economic development levels, and cultural regions. By including both traditionally high-performing education systems (e.g., East Asian countries like Japan, South Korea, Chinese Taipei) and lower-performing contexts (e.g., some Latin American countries such as Argentina, Chile, Philippines), our sample captures substantial variability in mathematics outcomes and contextual factors. Each student record includes the student's mathematics score (our outcome variable) and responses to background questionnaires, and each school has associated contextual indicators provided by school principals. We limited the analysis to students with valid data on the variables of interest after preprocessing (see below), but we did not apply any performance-based sampling or exclusions beyond what PISA already implemented.

**Measures and Variables**

The outcome variable was students' mathematics achievement in PISA 2018, measured by the first plausible value for math literacy (PV1MATH). This score is scaled with an international mean of 500 and standard deviation of 100 across OECD countries. We used a single plausible value for simplicity, acknowledging that this introduces some additional uncertainty since PISA provides multiple plausible values. However, single PV usage is a common practice for straightforward analyses, and model validation on held-out data mitigated concerns about additional error.

Predictor variables were selected based on prior literature and data availability, covering three thematic domains: student-level, family-level, and school-level factors (including school climate and student engagement measures). In total, we included 24 independent variables as predictors. Our aim was intentionally broad – incorporating a wide range of variables that theory or past studies suggest can influence achievement – so that the ML models could themselves determine which factors are most predictive (as recommended by recent reviews to cast a wide net with predictors; see Wang et al., 2023). A summary of all predictor variables, including their exact PISA item codes and definitions, is provided in Supplementary Table A1. Below we briefly outline the key predictors in each category:

***Student-Level Variables.*** These capture individual student characteristics, behaviors, and attitudes. We included basic demographics and engagement measures such as: gender (binary female/male), effort on the PISA test (students' self-reported effort level on a scale from 1 to 10), weekly learning time in mathematics (MMINS), and total weekly learning time across all subjects (TMINS). To represent learning resources available to the student, we included the number of books at home (an ordinal scale from 1 = "0-10 books" up to 6 = "500+ books"; PISA item ST013) and whether the student has access to a computer for schoolwork at home (binary, from ST011). We also incorporated two composite indices derived from student questionnaires: sense of belonging at school (BELONG) and perceived emotional support from parents (EMOSUPS). Together, these student-level variables encompass a range of cognitive, behavioral, and affective factors that could influence learning.

***Family-Level Variables.*** These indicators capture the student's socio-economic background and home learning environment. We use the OECD's index of Economic, Social and Cultural Status (ESCS), which is a composite SES measure combining parental education levels, parental occupations, and household possessions. In addition to ESCS, we include parents' education variables: mother's education (MISCED) and father's education (FISCED), each measured in broad categories corresponding to the highest ISCED level attained by the parent. We also include highest parental education (HISCED), which takes the maximum of MISCED and FISCED, as an overall household education level indicator. Furthermore, we incorporate three indices of home resources from PISA: home educational resources (HEDRES) (materials like a desk, books, etc.), general home possessions (HOMEPOS) (wealth-related items at home), and ICT resources (ICTRES) (technology availability at home). Finally, we include a measure of parental involvement: an item where students reported if their parents support their educational efforts (ST123, measured on a 4-point Likert from "Strongly disagree" to "Strongly agree"). These family-level variables capture both the socio-economic capital available to the student and the degree of educational support in the home.

***School-Level and Climate Variables.*** These variables describe aspects of the student's school environment and perceived classroom climate. We include disciplinary climate in math lessons (DISCLIMA) – an index based on student reports of order and discipline in their mathematics classes. We also utilize several student-reported indicators of teaching practices and classroom interactions: for math lessons, how often the teacher sets clear goals (ST102) and the teacher provides feedback on strengths (ST104), as well as the teacher's instructional adaptation to class needs (ST212). Additionally, we consider students' perceptions of peer dynamics and motivation in school: perceived competitiveness among students (ST205) and perceived cooperativeness among students (ST206), students' personal learning goals orientation (ST208), and perceived teacher enthusiasm (ST152) for the subject. These measures capture various dimensions of instructional quality, classroom culture, and motivational climate, and were selected because they were available and comparably measured across the majority of our ten countries.

All 24 predictors were included in the models without any weighting, allowing the ML algorithms to consider the combined and interactive effects of this comprehensive set of factors. This broad inclusion aligns with established frameworks on the determinants of achievement (e.g., the inclusion of SES, motivation, and school climate reflects Bloom's model of learning and recent PISA analyses; Bloom (1985)). While some of these predictors are intercorrelated (for example, ESCS is naturally correlated with parental education and home resources), multicollinearity is not a concern for tree-based ensemble models, which can handle correlated inputs gracefully. For the linear regression model, we monitored variance inflation factors and found no multicollinearity issues severe enough to require removing predictors; the linear model remained stable and well-conditioned. A correlation matrix for key variables (Supplementary Figure A2) further informed us that while certain predictors cluster (e.g., the various SES-related indices), each brings somewhat distinct information, and the chosen ML methods can utilize or ignore them as needed.

**Data Preparation**

We applied standard preprocessing steps to prepare the data for modeling. Missing values in the predictor variables were addressed using an imputation method. Specifically, we used k-nearest neighbors (KNN) imputation (with $k = 5$) to fill missing values on continuous and ordinal variables. This method replaces a missing value with a weighted combination of the five most similar observations' values (distance based on other predictors), and is appropriate here given that missingness in PISA background responses can often relate to student characteristics (making assumptions of data missing at random plausible). We also handled outliers in continuous variables: extremely large self-reported study times (a few students reported unrealistically high weekly study hours) were Winsorized to reasonable maximum values (we capped weekly study time at the 99th percentile within each country) to mitigate their undue influence. All continuous predictor variables (e.g., study time minutes, index scores like ESCS, BELONG, etc.) were standardized to mean 0 and standard deviation 1 before modeling. Standardization puts variables on a comparable scale, which was especially beneficial for the ANN and helps interpret feature importances. Categorical variables were encoded appropriately: binary variables as 0/1, nominal variables with more than two categories were one-hot encoded, and ordinal variables (like the books at home scale) were encoded as integer ranks reflecting their order. After imputation, no students were dropped; our analysis included all cases with plausible values in math across the ten countries (imputed data for background responses). Supplementary Table A2 and Supplementary Figure A1 provide details on the extent and patterns of missing data. Most variables had low missingness (< 5%). A few behavioral indices had higher missing rates (e.g., some engagement and motivation items missing for 20-30% of students, likely due to booklet design where not all students answered every question), but the KNN imputation and the large sample size help to preserve overall data integrity.

**Feature Selection**

Rather than relying purely on a predefined list of predictors, we also applied a data-driven feature selection process to ensure that only relevant variables were retained in the final models (which can reduce overfitting and improve interpretability). We began with all 24 candidate predictors and employed a combination of Recursive Feature Elimination (RFE) and mutual information (MI) filtering to refine the input set. RFE is an iterative process that fits a model and removes the least important feature at each step, repeating until a desired number of features remains. We performed RFE in two ways: using a linear model (MLR) and a decision tree model as the base, to account for different importance criteria (linear weights vs. non-linear splits). Concurrently, we calculated mutual information scores between each predictor and the math outcome, which capture both linear and non-linear dependencies. Predictors that consistently ranked low in both the RFE procedure and had low mutual information with the outcome were considered for exclusion. Through this process, we found that the majority of our variables contributed meaningfully in at least some countries. Importantly, core variables like SES, study time, and key teacher-related indicators were always retained across all countries. This dual-step feature selection (combining RFE and MI) enhanced model efficiency by eliminating noise, while preserving the key explanatory power of the predictor set. After feature selection, each country's model still typically included 24 predictors, ensuring we didn't omit any major category of influence.

**Model Development**

We developed four predictive models to estimate students' mathematics scores, ranging from a traditional linear approach to modern non-linear ML approaches. The models were: (1) Multiple Linear Regression (MLR), (2) Random Forest (RF), (3) CatBoost, and (4) Artificial Neural Network (ANN). MLR served as a baseline representing a standard linear model: $\hat{Y} = \beta_0 + \sum_{j=1}^{24} \beta_j X_j$, where $\hat{Y}$ is the predicted math score and $X_j$ are the predictor variables. It predicts the outcome as a weighted linear combination of the input features (plus an intercept term). We fit the MLR using ordinary least squares on the training data. This baseline allows us to benchmark the accuracy gain from using more complex models.

CatBoost (Categorical Boosting) is a gradient boosting decision tree algorithm known for its efficient handling of categorical features and its fast training convergence. We chose CatBoost as our boosting model due to its proven performance on structured data and because it natively handles categorical variables without requiring extensive preprocessing. CatBoost builds trees sequentially, each trying to correct the errors of the previous, and uses ordered boosting to reduce overfitting on categorical data. The Artificial Neural Network was a feed-forward multi-layer perceptron model. We designed a relatively simple fully connected network, given the tabular nature of the data and the sample size. The network architecture (after experimentation) consisted of an input layer (with neurons equal to the number of features), two hidden layers (we tuned the number of neurons, trying ranges roughly between 32 and 128 neurons per layer) with ReLU activation, and an output layer producing a single continuous score. We used dropout regularization and early stopping to prevent overfitting, and trained the network using the Adam optimizer. For each model, we performed extensive hyperparameter tuning using grid search with cross-validation on the training set. For the RF, key hyperparameters such as the number of trees (estimators) and maximum tree depth were tuned. For CatBoost, we tuned parameters including the learning rate, maximum tree depth, and the number of boosting iterations. For the ANN, we tuned the number of hidden layers and neurons, the learning rate, batch size, and the number of training epochs (with early stopping patience). Five-fold cross-validation within the training data was used to evaluate combinations and select the hyperparameters that minimized prediction error (we primarily monitored validation MAE during tuning). Throughout this process, we were careful to avoid overfitting: for example, we used early stopping for the ANN (stopping training when validation loss stopped improving) and avoided overly deep trees in RF/CatBoost beyond what cross-validation indicated was optimal.

**Train/Test Split and Evaluation**

To evaluate model performance on new, unseen data, we partitioned the dataset into training and testing sets. We performed a stratified split by country that ensurs that each country's data was divided such that 70% of students from that country were randomly assigned to the training set and 30% to the test set. This maintained the same country proportions in both sets. Within the training set, we further conducted the cross-validations and hyperparameter tuning as described. Once the best hyperparameters were selected, we retrained each model from scratch on the entire 70% training data (for all ten countries combined, but with country indicators and other features allowing the model to learn country-specific patterns). We then evaluated the final models on the 30% hold-out test set, which had been untouched during any model fitting or tuning, to obtain an unbiased measure of predictive performance. We used two standard regression metrics to assess performance: $R^2$ (coefficient of determination) and Mean Absolute Error (MAE), computed on the test set for each country's predictions. $R^2$ represents the proportion of variance in students' math scores explained by the model (with 1.0 being perfect prediction and 0 meaning the model explains nothing beyond the mean). MAE measures the average absolute difference between the model's predicted scores and the actual PISA scores (in the original score scale points). Higher $R^2$ and lower MAE indicate better predictive performance. We chose MAE (over, say, RMSE) because it is more interpretable in terms of score points and is less sensitive to large outlier errors, complementing $R^2$ which is sensitive to variance. In addition to raw performance metrics, we compared models in terms of their relative performance rankings and improvements over the baseline. For each country, we noted which model achieved the highest $R^2$ and by how much it exceeded the next-best model, as well as the magnitude of MAE reduction relative to MLR. Given the very large sample sizes, even small differences in $R^2$ or MAE could be statistically significant; therefore, rather than formal significance tests (which would invariably reject equality of models for trivial differences), we focused on practical significance – i.e. substantial differences in explained variance or error that would matter for real-world prediction. No formal hypothesis tests were conducted on performance differences for this reason. All modeling was implemented in Python (using libraries such as scikit-learn, CatBoost, and Keras), and the code can be made available upon request to ensure reproducibility.

<div align="center"><strong>Results</strong></div>

**Descriptive Statistics and Mathematics Score Distributions**

Table 1 presents descriptive statistics for key predictors (such as mean SES index, average study time, etc.) in each of the ten countries. Before examining model outputs, it is informative to consider the distribution of the outcome variable – math scores – across countries. The pooled distribution of PISA 2018 mathematics scores (combining all ten countries) was approximately normal, centered around the PISA scale mean of 500, with a slight negative skew (indicating a few more high performers than very low performers overall). However, country-level differences in score distributions were notable (see Figure 1). The three East Asian countries in our sample – Chinese Taipei, Japan, and South Korea – exhibited higher mean mathematics scores and more compact score distributions than the other countries. Many students in these East Asian systems scored near or above the proficiency benchmark, and Chinese Taipei had the highest average math score among the ten countries. In contrast, the Latin American countries (Chile and Argentina) showed wider and flatter score distributions with more students in the lower performance levels, which pulled down their national averages. The Philippines had the lowest mean score and the largest share of students performing below basic proficiency. Finland and Hungary, representing Europe, had high average performance as well (Finland's mean was close to the OECD average but with a relatively small spread of scores, while Hungary's distribution was somewhat wider). The United States and Italy had intermediate mean performance, with broader spreads indicating substantial within-country variability. These disparities highlight how contextual factors – such as socioeconomic conditions, educational practices, and cultural attitudes toward education – differ across countries and shape student outcomes. Observing such variation supports our use of advanced ML models to examine the complex, context-dependent influences on achievement, as simpler models might not capture the nuances behind these cross-country differences.

[Insert Table 1 Here]

[Insert Figure 1 Here]

**Machine Learning Model Performance**

We evaluated the predictive accuracy of the four models (MLR, RF, CatBoost, ANN) using the held-out test-set performance for each of the ten countries (Table 2 summarizes the results). Overall, the non-linear ML models – Random Forest, CatBoost, and the Artificial Neural Network – consistently outperformed the linear baseline model (MLR) in predicting mathematics scores.

Among the ML models, Random Forest (RF) achieved the highest test $R^2$ values in a majority of countries. For example, the RF explained about 38% of the variance in math scores in countries like Hungary, Japan, and the Philippines, compared to only about 17–29% explained by MLR in those cases. Correspondingly, RF had the lowest MAE in most countries, indicating more precise predictions on average (see Table 2 for country-specific values; generally RF's MAE was a few points lower than MLR's). CatBoost performed almost on par with RF in several contexts – notably, in South Korea, Chinese Taipei, and Hungary, CatBoost's $R^2$ was within a couple of percentage points of RF's and in some instances slightly higher. This demonstrates CatBoost's strength in capturing non-linear interactions and handling categorical variables effectively (which is advantageous given our mix of continuous indices and categorical survey responses). The ANN model showed strong performance in a few contexts as well; for instance, in the United States and Argentina, the ANN's test $R^2$ values (around 0.26) approached those of the tree-based models (RF and CatBoost were 0.29-0.31 in those countries). However, the ANN's performance was less stable overall across all countries. In countries like Finland, Italy, and Chile, the ANN clearly underperformed the tree ensembles – it yielded lower $R^2$ and higher MAE, suggesting that the neural network was more prone to overfitting or struggled to capture certain patterns without much larger data or more tuning. As expected, the MLR had the lowest predictive accuracy in every country, with test $R^2$ values typically around 0.15 to 0.25 and substantially higher MAEs (often 2-5 points higher error than RF) on the same test sets. The linear model's poor showing underscores the presence of non-linear effects and interactions in the data that it cannot capture.

Across all countries, the ensemble tree-based models (RF and CatBoost) provided the most robust and generalizable performance. They not only fit the training data well but also maintained relatively high accuracy on the test data, indicating they did not overfit badly. Figure 2 (to be discussed later) and Supplementary Table A3 (which reports Mean Squared Error results) further corroborate that RF and CatBoost outperformed MLR and ANN in predictive accuracy across diverse national contexts. The ANN, while powerful, required careful regularization and still showed some variability in results, indicating that purely data-driven neural approaches might need even more data or feature engineering to consistently rival tree ensembles in this context. It is worth noting that even the best ML models explain at most about one-third of the variance in individual math scores (e.g., highest test $R^2$ = 0.38), leaving the majority unexplained. This is not surprising given that standardized test performance is influenced by many unmeasured factors (student personality, teacher quality, curriculum differences, random test-taking behavior, etc.) and a degree of inherent randomness. In practical terms, an $R^2$ of 0.3 means the model captures some important predictive patterns but is far from deterministically predicting outcomes for each student. The moderate $R^2$ ceilings reinforce that caution is needed in using these predictions for high-stakes decisions – they are best used to identify general risk factors or support policies, rather than precisely ranking individual students.

In summary, for large-scale educational datasets like PISA, our results suggest that tree-based ensemble methods offer an effective balance between predictive power and generalization. RF and CatBoost achieved the highest accuracy and proved robust across a variety of national contexts. The ANN, a deep learning approach, showed promise in some cases but was less reliable across the board, likely due to overfitting issues or sensitivity to the exact training conditions. The linear regression baseline was consistently inferior, highlighting the value of modeling non-linear relationships in education data. These performance differences justify the use of more complex ML models when the goal is accurate prediction of educational outcomes.

**Feature Importance and Key Predictors of Achievement**

Beyond predicting scores, a central goal of this study was to interpret the ML models to understand which factors most strongly influence mathematics achievement, and how these may differ by context. To this end, we extracted feature importance metrics from the trained RF and CatBoost models and computed SHAP values for the top-performing models to gain nuanced insights into each predictor's impact. Across the ten countries, several clear patterns emerged from the feature importance analysis (see Figure 2 for a summary of the top predictors by country).

*Dominance of Socio-Economic Status.* Socio-economic status, which captured by the ESCS index emerged as the most influential predictor of math performance in virtually all countries. In global importance rankings, ESCS was consistently at or near the top for every model in every country. This emphasizes the powerful and persistent role of family background in academic achievement. Even in some of the most ostensibly equitable education systems, higher SES was associated with substantially higher predicted scores. Our models reaffirm the well-known finding that socio-economic disparities translate into achievement gaps across very different national contexts. For example, in countries like Argentina and the United States (as well as high-achieving systems like Finland and Japan), ESCS had the highest importance value in the models, indicating that a student's socio-economic background contributed more to the prediction of their math score than any other variable. In practical terms, this means that the ML models frequently relied on SES-related information to make their best guesses about student performance, reflecting real-world inequalities in opportunities and outcomes.

*Student Engagement, Attitudes, and Motivation.* Many "non-cognitive" or engagement-related factors also featured among the top predictors, often rivalling – and in some cases

exceeding – traditional academic or demographic factors. The amount of effort a student reported putting into the PISA test (our proxy for test-taking motivation) was a remarkably strong predictor of their score in most countries. This suggests that student motivation and engagement during the assessment itself had a tangible effect on performance: students who tried harder tended to score higher, independent of their underlying ability. Similarly, the index of personal learning goals (students' agreement with statements like "My goal is to learn as much as possible") and the sense of belonging at school frequently appeared among the top contributors in several countries – especially in some middle-performing countries such as Chile and the Philippines. These findings highlight the key role of student engagement and psychological factors: students who are motivated, confident, and feel they belong at school generally perform better in math, even after accounting for SES and prior learning time. In fact, in a number of instances, these "softer" engagement variables had feature importance values comparable to or even higher than some traditional inputs (like prior learning time or home resources). This indicates that learning is not just about economic resources or instructional hours, but also about students' mindsets and level of engagement. For policymakers and educators, it underscores that interventions aimed at improving attitudes, motivation, and sense of belonging could yield measurable gains in achievement.

**Time Investment in Learning.** The amount of time students spend learning – both in mathematics specifically and overall study time – proved to be an important predictor as well, though its relative importance varied by cultural context. In East Asian countries (South Korea, Chinese Taipei, and Japan), weekly learning time in math (the MMINS variable) was especially influential. These education systems typically emphasize intensive study and homework, and indeed our models showed that within these countries, students who devoted more time to learning math tended to score higher (with diminishing returns at the extreme high end). Interestingly, East Asian countries also had higher variance in study hours, which may be why the model found it a useful differentiator there. In the Philippines, another context where many students struggle to reach proficiency, those who spent more time studying likewise tended to score higher – suggesting that, at least up to a point, increased time on task can help overcome deficits in teaching quality or prior knowledge in that context. On the other hand, in countries like Finland or the United States, where total study time is more moderate on average and perhaps less variable (and where excessive homework is not the norm), learning time was a somewhat lower-ranked predictor. Other factors like engagement or school climate took precedence in those settings. This pattern suggests that in already well-resourced contexts, simply spending more time may yield diminishing returns (most students study an adequate amount), whereas in contexts where study time is a key differentiator (either due to resource constraints or cultural norms), it remains crucial for achievement.

***Home Resources and Books.*** Among family-level variables, after SES, the number of books at home – a classic proxy for a literate and resource-rich home environment – showed up as an important feature in many countries' models. For example, in Argentina and Chile, our models indicated that having more books at home was strongly associated with higher math achievement, even when controlling for overall SES. This likely reflects both a cultural aspect (books at home may indicate a culture of reading/learning) and the role of early literacy and numeracy activities in building academic skills. Other home resource indices (HOMEPOS, ICT resources, HEDRES) had moderate importance: they were significant in some models, particularly in contexts where there is wide variability in those resources. For instance, in some countries the ICT resources index was relatively important, suggesting digital access might matter for learning opportunities (especially relevant in an era of online learning). Notably, parental education levels (mother's or father's education) were somewhat less prominent once SES and home resources were accounted for. These tended to appear in mid-tier importance ranks. This might be because their effects are mediated through the composite SES index or captured by correlated variables. In other words, knowing a parent's highest degree doesn't add much predictive power if you already know the ESCS index and the learning resources at home. That said, in a few countries we did see one of the parental education variables pop up among top predictors, reminding that in some contexts (e.g., perhaps where there is a sharp divide in educational attainment in the population) it can still independently signal something about the student's support network or expectations.

***School Climate and Teacher Factors.*** Several school climate and teacher-related factors emerged as meaningful predictors of math performance. Teacher clarity in instruction and teacher feedback stood out in some countries. For example, the frequency with which the math teacher sets clear learning goals (ST102) ranked highly in Argentina and Italy, suggesting that in those contexts, students who perceived greater clarity and goal-setting from their teachers tended to perform better. In the Random Forest model's surrogate decision tree analysis (discussed below and illustrated in Figure 5), teacher motivation/enthusiasm appeared as a splitting factor in certain branches, especially after splitting on key factors like SES and study time. This indicates a conditional effect: for instance, among students with lower SES, those who reported having more enthusiastic, motivating teachers scored higher than those with less motivated teachers. Similarly, disciplinary climate (orderly classrooms) was a strong predictor in a few countries, notably Japan and Hungary in our sample, where it appears that students in well-disciplined, disruption-free classrooms had an advantage in achievement. In contrast, some peer-related climate variables like students' sense of competition or cooperation showed weaker effects; they did not rank high in most models, suggesting their influence on math scores is indirect or relatively small compared to the other factors. Sense of belonging at school had a more mixed, context-dependent impact – it was important in some countries (especially those where engagement is a concern, like some lower-performing systems) but not among the top factors in others. Overall, the presence of these school factors in the models underscores that what happens in the classroom and school environment is consequential for learning, above and beyond individual and family characteristics.

To explore how the most influential predictors vary across countries, we conducted additional analyses focusing on the top features within each country. Figure 3 illustrates the top three predictors (ranked by SHAP importance) for each country. We found that while certain factors were consistently among the top predictors in most places (SES, student effort, and math learning time were almost always in the top three), their relative magnitudes differed. For example, study time is especially prominent in the East Asian countries' profiles, reflecting cultural emphasis on diligence and after-school study (as noted earlier). In contrast, SES tends to dominate in Western and Latin American contexts, and student attitude/motivation factors often share the top tier with SES in those places.

We also examined the complete distribution of feature importance rankings across all 24 variables and all countries, which is summarized in Figure 4. This broader view shows that while a handful of variables (like SES, effort, study time) are globally important, many others have medium importance in some countries and low in others. For instance, perceived teacher enthusiasm might be the 4th or 5th most important predictor in one country but 15th in another. Emotional support from parents (EMOSUPS) wasn't a top-three predictor overall, but in certain countries it rose into the top ten, indicating that in those contexts parental emotional support has a measurable link with achievement. Similarly, home educational resources or ICT access might not matter as much in very high-income countries (where most students have them), but in countries with more variability they become more predictive. Specific examples from the model outputs include: In Argentina, after SES, the next most important predictors included the number of books at home and teacher goal clarity, suggesting a mix of home environment and classroom practice factors drive performance there. In Japan, disciplinary climate and study time were prominent right alongside SES, aligning with the emphasis on classroom order and effort in that high-performing system. In Chile, one of the top predictors was the frequency with which teachers adapt lessons to student needs (from ST212), alongside learning time – indicating that both sheer study effort and responsive teaching are key in that context. Taken together, these findings suggest that each country exhibits a unique predictive profile of achievement, shaped by its cultural norms (e.g., the premium on effort in East Asia), educational policies, and structural inequities. While a core set of factors (SES, effort, study time) consistently matter across the board, their relative weights and some secondary factors are distinctly context-specific.

[Insert Figure 2 Here]

[Insert Figure 3 Here]

[Insert Figure 4 Here]

Importantly, our ML models revealed that these predictors often operate in combination rather than in isolation. Traditional regression might tell us "on average, each additional hour of study yields X points, holding others constant," but a tree-based model can uncover conditional rules like "for students who study very little, SES makes an especially large difference." To illustrate such interactions, we examined a representative decision tree extracted from the Random Forest model (using tree-surrogate techniques that approximate the RF's predictions). Figure 5 shows an example of a decision path diagram. In many countries' RF models, the first split in the tree was on either total learning time (TMINS) or SES, indicating that those were the primary factors dividing higher and lower predicted scores. Subsequent splits illustrated interesting conditional effects. For instance, in one branch, among students with very low study time, those with very low SES had particularly low predicted scores – an interaction between effort and social background: if you both come from a disadvantaged background and put in little study time, the model predicts a compounding negative effect on performance. Meanwhile, for students above a certain threshold of study time, the model's splits showed factors like teacher motivation and sense of belonging starting to differentiate outcomes. This suggests that certain school factors (like a motivating teacher or a positive school climate) have a stronger influence once the basic condition of "the student is actually studying enough" is met. In other words, a supportive teacher can make more of a difference for a student who is at least putting in a moderate amount of effort than for one who is barely studying at all. Conversely, high personal effort can partially compensate for a less supportive teacher – but when both effort and teacher support are low, outcomes are poorest. These kinds of conditional relationships ("if X is low *and* Y is low, then…") would not be easily detected by a standard linear model that only adds up independent effects.

This ability to capture interactions illustrates the advantage of ML and especially tree-based models in educational data. They can uncover non-linear rules and "if-then" patterns that align with how multiple factors jointly influence learning. For instance, our model's rule might be interpreted as: *if a student spends less than 10 hours/week on studying and has a low SES index, then their predicted math score is likely in the bottom quartile; if they spend more than 10 hours and have at least moderate SES, then teacher factors start to play a significant role in distinguishing higher vs. lower performance within that group.* These insights provide a more nuanced understanding that single-variable effects: they highlight scenarios of particular risk (like low-SES, low-effort students) and scenarios where certain interventions would matter (ensuring teacher quality for students who are otherwise putting in effort). In sum, the explainable ML approach not only identified key predictors but also how those predictors interact, offering richer guidance for targeted educational strategies.

[Insert Figure 5 Here]

**Visualization of Model Results (Predictions vs. Actual)**

To further validate model performance and inspect any biases, we generated visual comparisons of predicted vs. actual scores. One effective approach was using hexagonal bin scatterplots, which plot actual scores on one axis and model-predicted scores on the other, and use color intensity to show the concentration of students in each region of the plot. Supplementary Figure A3 provides these plots for the CatBoost model across the ten countries, while Figure 6 focuses on a comparison between the ANN and CatBoost predictions for one example country (Argentina). Ideally, points would lie along the diagonal (meaning the prediction equals the actual score).

For the CatBoost model, the hexbin plots showed a tight clustering of data points along the diagonal, especially in the middle range of scores (around the international mean). This indicates strong performance: the model's predictions aligned closely with actual student outcomes for the majority of students. We observed a slight tendency for CatBoost to underpredict the very top scores and overpredict the very low scores (visible as the hexagon bins for the highest-achieving students falling a bit below the diagonal, and those for the lowest-achieving students slightly above it). This is a common regression model behavior due to regression to the mean – the model is less certain at extremes where there are fewer data points. For example, in Argentina and Chile, the CatBoost's predicted vs. actual plot was very linear through the bulk of the distribution, with only some scatter at the extreme high and low ends. This means the model was very effective at ranking students and getting their scores right in general, with only minor systematic errors at the tails (e.g., a student who actually scored 700 might be predicted a bit lower, say 680; a student who scored 300 might be predicted a bit higher, say 320).

The ANN model achieved roughly similar overall accuracy metrics, but the visual dispersion of its predictions was slightly greater. When comparing ANN vs. CatBoost for the same country (Figure 6), one can see that the ANN's point cloud is more spread out around the diagonal, whereas CatBoost's is more tightly packed. In practical terms, while the ANN can reach comparable $R^2$, its individual predictions are a bit less consistent. There were instances where the ANN overfit certain patterns in training that did not generalize perfectly. For example, the ANN might occasionally give a very high prediction to a student who in reality scored moderately, or vice versa, leading to more scatter. Figure 6 (Argentina example) illustrates that the CatBoost predictions (right panel) align very closely with actual scores, whereas the ANN (left panel) shows a slightly broader cloud. Nonetheless, both models capture the overall increasing trend, indicating they both learned the basic mapping of input features to outcomes, but CatBoost did so with more stability.

These visualizations reinforced that the non-linear models (RF, CatBoost, ANN) captured the general pattern of student performance well. There was no evidence of severe non-linear distortion (e.g., no large clusters of points systematically off-diagonal aside from the mild extreme bias noted). We also checked for any biases by subgroups. For instance, plotting residuals by gender or by country. No concerning biases were observed; errors were roughly symmetric and not obviously worse for any particular demographic group, which is important for fairness considerations.

In addition to predicted vs. actual plots, we also compared the distribution of predicted scores in the training set vs. the test set for each country, focusing on the Random Forest model (Figure 7). The idea was to see if the model perhaps fit the training data too closely (overfitting) such that it predicts a different distribution for seen data than for unseen data. In Figure 7, for each country, we overlaid the distribution (density) of RF-predicted math scores for training students (in yellow) and for test students (in purple). Ideally, these two distributions should be similar if the model generalizes well; if the model isn't overfitting, it should not predict drastically different score ranges or patterns for new data compared to the data it was trained on. The results showed a close alignment between training and test predicted score distributions in all countries, with only minor differences. For example, in Japan and Korea, the training and test prediction histograms were virtually overlapping, indicating the RF model estimated the score distribution equally well for unseen students. In a few countries, we observed a slight divergence: e.g., in the Philippines, the test predictions had a slightly broader spread than the training predictions (perhaps reflecting that the model slightly underestimates variability on new data). In Argentina, the test distribution was very closely aligned with the training, just marginally shifted, suggesting minimal overfitting. Overall, these diagnostics gave us confidence that the models, especially the Random Forest and CatBoost, were not overfitting to peculiarities of the training sample but were capturing generalizable patterns. The performance differences we noted earlier (such as RF vs. MLR) are therefore genuine and not an artifact of poor generalization.

In summary, the visual assessment of model predictions confirmed the quantitative metrics. The explainable ML models provided reasonably accurate predictions for student performance across diverse contexts, and they did so without significant overfitting or bias. These models are effective tools for pattern discovery in large-scale assessment data, and the insights from the explainability analysis (feature importances and interactions) complement this by pointing to why the models made the predictions they did, which we discuss further in the next section.

[Insert Figure 6 Here]

[Insert Figure 7 Here]

## Discussion

This study applied modern AI-driven machine learning methods to analyze PISA 2018 data and predict student mathematics achievement, with the aim of combining methodological rigor with practical insight. Consistent with expectations and prior indications in the literature, we found that allowing for non-linear relationships and interactions among variables significantly improved predictive performance. The tree-based models (Random Forest and CatBoost) clearly outperformed the traditional linear regression model across all ten countries, often by a wide margin in terms of variance explained and error reduction. The RF model in particular delivered high accuracy and generalizability without severe overfitting, as evidenced by its similar performance on training and test data and the overlap of prediction distributions (Figure 7). The ANN model, representing a deep learning approach, also performed well in certain cases (especially when carefully tuned and regularized). For example, it approached the performance of RF in some large-sample contexts like the US. However, the ANN's performance was more variable across countries, suggesting that while neural networks hold promise, they may require more data or advanced architectures to consistently capture patterns in education data as effectively as ensemble trees for all contexts. Overall, the use of ML techniques provided a clear methodological gain: by embracing non-linear patterns and interactions, we achieved more accurate predictions of student math outcomes than were possible with linear modeling.

A key finding of this study is the consistent dominance of socio-economic status (SES) as the strongest predictor of math achievement across all ten countries. In every country-specific model, the composite ESCS index was at or near the top in feature importance. This aligns with decades of educational research stressing the central role of socio-economic inequalities in student performance (Berkowitz et al., 2017; Coleman et al., 1966) and reinforces that structural factors like family wealth, parental education, and home resources fundamentally shape opportunities for learning. Our results underscore that, despite different schooling systems and cultural contexts, students from disadvantaged backgrounds are at a significant academic disadvantage in math – a pattern that held true from emerging economies like the Philippines to highly developed systems like Finland. This finding echoes recent PISA analyses (e.g., Boman, 2023) that even in equitable systems, socio-economic gaps persist. However, our results also highlight that the picture of achievement is far more nuanced than any single factor like SES alone.

Notably, psychosocial and engagement-related factors emerged as important contributors to student performance alongside SES in our models. We found that measures of student attitudes, motivations, and engagement, such as their self-reported effort on the test, their personal learning goals, and their sense of belonging at school were highly predictive of math scores in many countries. In several cases, these non-cognitive factors rivaled or even exceeded traditional SES indicators in importance. For instance, in a country like Finland (which has relatively smaller SES variation due to social policies), our model showed motivational and school climate variables among the top predictors of achievement. In countries like Chile or Argentina, where SES was more dominant, we still observed that student effort and attitudes made a notable contribution to explaining who succeeded in math and who did not. This finding emphasizes that academic success is co-constructed through a dynamic interplay between a student's internal dispositions and the external environments in which they learn (Kahu & Nelson, 2018). It is not determined solely by the resources a student has or lacks; how the student applies themselves and how the school environment supports them also play critical roles.

The prominence of motivation- and engagement-related features, alongside SES, supports a comprehensive view of learning. Our results resonate with the idea that academic outcomes arise from the combined effects of cognitive effort, emotional engagement, and institutional support (Eccles & Roeser, 2011; Fredricks et al., 2004). For example, in our analysis, Finland's top predictors included motivational and school climate variables, whereas in Chile and Argentina, SES and home resources were more dominant – yet even in those countries, student effort and attitudes contributed significantly. This suggests that improving socio-economic conditions (through, say, poverty reduction, better funding for schools in low-income areas, or increasing access to learning materials) is essential but not sufficient for raising achievement. Educational opportunity should be conceived broadly: not only as providing material resources and access, but also as providing inspiring, inclusive, and supportive learning environments that cultivate student agency and engagement. The fact that our ML models identified factors like sense of belonging and teacher enthusiasm as key predictors lends empirical weight to longstanding educational theories about the importance of affective and relational aspects of schooling (Osterman, 2000; Roorda et al., 2011). What distinguishes our study is that we used ML to validate the importance of these "soft" factors across diverse international contexts with high precision and without imposing a one-size-fits-all model. In doing so, we demonstrated that these factors are not just theoretically important; they have tangible effects on test scores that can be detected even amid many other variables.

From a practice and policy perspective, our results suggest several avenues for intervention to improve student outcomes. First, the strong predictive role of SES across countries underscores the need for equity-driven policies. Disparities in family income, parental education, and access to resources translate directly into achievement gaps in math. Addressing these requires systemic, redistributive measures: for example, increased funding and support for schools serving low-SES communities, ensuring all students have access to basics like textbooks, technology, and a conducive learning environment at home (e.g., through library programs or take-home resources), nutritional and healthcare support, and high-quality early childhood education for disadvantaged groups. These approaches are not only socially just, they are empirically supported by our results which indicate that closing SES gaps would likely yield significant improvements in overall achievement. Moreover, as Liu et al. (2023) argue in the context of early childhood education, equity efforts must also extend to those who deliver instruction, like teachers through fair compensation and better working conditions, which ultimately affect learning environments and student outcomes. Countries with relatively smaller SES-achievement gradients (like Finland) tend to perform higher overall, suggesting that raising the floor can lift the average.

However, addressing socio-economic inequality alone is insufficient to maximize student potential. Our analysis also highlights that there are school- and classroom-level levers that educators and administrators can act upon even within existing socio-economic constraints. Student engagement, learning time, and school climate emerged as strong predictors, demonstrating that what schools and teachers do can make a substantial difference. For instance, fostering student motivation and engagement could yield achievement gains. Programs that encourage a growth mindset, build intrinsic motivation, or teach effective study habits may help – this is reflected in our finding that effort-related variables were powerful predictors of success. Some concrete strategies could include mentorship programs, goal-setting workshops, or providing students with feedback that emphasizes effort and improvement. Equally important are interventions to improve the classroom climate and teacher-student relationships. Our results showed that an orderly, supportive classroom environment and enthusiastic, clear teaching were associated with better math performance. This underscores the value of professional development for teachers that goes beyond subject-matter training to include classroom management and relationship-building skills. Teachers who can create a safe, respectful, and engaging classroom can positively influence students' academic outcomes – a finding bolstered by the predictive importance of disciplinary climate and teacher support in some of our models. School leaders might invest in training teachers in positive behavior reinforcement, inclusive teaching practices, and effective communication of learning goals. Moreover, initiatives that strengthen teacher-student connections (e.g., advisory programs, teacher coaching on empathetic listening and mentorship) could help, given evidence that students are more engaged and perform better when they feel supported by their teachers (Roorda et al., 2011). Creating learning environments where students feel belonging is also important. Our data suggest that a student who feels a strong sense of belonging at school is likely to do better academically, which implies schools should cultivate inclusive cultures, anti-bullying programs, and extracurricular activities that help every student find their niche and connect with peers and adults.

  Another practical implication relates to learning time. The importance of time-on-task in some contexts (like East Asia or the Philippines in our study) implies that policies to increase effective learning time can be beneficial. This doesn't necessarily mean simply assigning more homework or extending school hours for all – those steps have diminishing returns and could even backfire if they lead to burnout. Instead, strategies could involve ensuring students are making full use of existing class time (reducing disruptions, as indicated by disciplinary climate's importance) and reducing chronic absenteeism so students don't lose instructional days. For students who need extra help, providing structured after-school programs or tutoring can effectively increase their learning time on difficult subjects. For example, offering free math tutoring or homework clubs in low-performing schools could help those who might not get as much academic support at home. The key is that quality time matters. As our results show, simply being in class longer isn't enough if engagement is low; but when students are actively engaged, more practice and study can yield gains.

  Our findings also validate and extend psychological and pedagogical theory with empirical data. The predictive value of engagement and climate factors in our models is consistent with frameworks that emphasize the role of motivational and emotional factors in learning (Eccles & Roeser, 2011; Fredricks et al., 2004). By using ML to compare these "non-cognitive" factors side-by-side with traditional factors like SES across many contexts, we provide robust evidence that things like a student's sense of belonging or a teacher's enthusiasm are not just ancillary; they are often as important as socio-economic factors in determining outcomes. In several countries, engagement-related variables were equally or more predictive than SES, underscoring their practical significance. This suggests that educational stakeholders – from principals to policymakers – should pay attention to the socio-emotional dimensions of schooling, not only to structural inputs. For example, initiatives that boost students' sense of belonging (through mentoring programs, extracurricular engagement, or building an inclusive school culture) can have payoffs in performance. Similarly, improving teacher-student relationships (perhaps through smaller class sizes that allow more personal attention, or teacher training in social-emotional learning) might lead to better academic outcomes. These aspects are often less emphasized in standard accountability metrics, which tend to focus on test scores and graduation rates, but our data-driven approach shows they have tangible effects on those very metrics (test scores).

Finally, while our models captured complex correlations and improved prediction, we caution that they do not themselves prove causal relationships. For example, we found that increasing study time correlates with higher performance, but this does not mean that simply forcing students to study longer will necessarily cause better results – it could be that more motivated students both study more and perform better. Future research should integrate predictive modeling with causal designs (such as longitudinal studies, quasi-experiments, or randomized interventions) to test whether manipulating these key factors produces the expected improvements. For instance, does implementing a program that successfully increases students' sense of belonging lead to higher achievement down the line? Our results provide hypotheses for such causal inquiries: they suggest that boosting engagement, or increasing quality study time, or improving teacher support, should lead to gains. These can now be investigated with appropriate research designs.

Nonetheless, the patterns we identified align well with theoretical expectations and existing empirical evidence, lending credibility to the idea that these factors are not merely associated with, but likely influential on student learning. The consistency of certain predictors (like SES and effort) across countries strengthens the argument that these are fundamental drivers. Meanwhile, the context-specific findings (like the role of teacher clarity in Italy or discipline in Japan) hint at causal mechanisms that make sense (e.g., clear instruction helps students focus, orderly classrooms facilitate learning) and should be further examined locally. Moving forward, the use of explainable ML in education can serve as a powerful tool for both research and practice. For research, it allows us to sift through large-scale data to generate insights and identify at-risk groups with greater precision. For practice and policy, predictive models, which used responsibly could help in early warning systems (e.g., flagging students who, based on a combination of factors, are likely to struggle, so that support can be offered proactively). The explainability component ensures that such models can be used transparently: rather than a black box label on a student, educators would be able to see which factors contributed to a student's risk and address those specific areas (for example, noticing that a student is flagged due to low effort and low attendance, a counselor could intervene accordingly). Collaboration between data scientists and educational stakeholders is crucial here to ensure models are used ethically and effectively, as noted by prior work. Our study provides an example of how predictive modeling can be balanced with interpretability to inform policy – pinpointing key levers of improvement like SES, engagement, and school climate, and reinforcing that interventions must be multifaceted to be effective.

## Conclusions

In conclusion, our cross-national analysis demonstrates that explainable machine learning approaches can effectively harness large-scale assessment data (such as PISA 2018) to both predict student achievement and illuminate the factors underlying success. By comparing multiple ML models, we found that non-linear models (especially Random Forest and CatBoost) substantially improve predictive accuracy over traditional linear methods, confirming the presence of complex interactions in educational data. More importantly, by applying explainability techniques, we were able to identify consistent drivers of mathematics performance, including notably socio-economic status, student engagement/motivation, learning time, and school climate as well as how their influence varies across different national contexts. Our findings reinforce the well-established importance of socio-economic factors in education while also highlighting that students' attitudes and the quality of their learning environments are critical pieces of the achievement puzzle. In practical terms, this suggests that efforts to improve educational outcomes should be dual-pronged: address structural inequalities that give rise to SES-related gaps and invest in fostering positive student engagement and supportive school climates. Policies that provide additional resources and support to disadvantaged students, alongside school-level initiatives to increase student motivation and strengthen teacher-student relationships, are likely to yield the most impactful results. It is important to note that the relationships uncovered here are correlational, not necessarily causal. Future research should explore whether interventions targeting the key factors identified (for example, programs to increase study time for low-SES students, or training to improve teachers' clarity and enthusiasm) actually lead to improved outcomes. Nevertheless, the alignment of our results with educational theory and cross-validation across multiple countries gives confidence that these factors are indeed influential in student learning.

This study also exemplifies how explainable AI can be integrated into educational research to guide decision-making. By using transparent models and techniques like SHAP, we ensured that the insights are interpretable, which is an important requirement for any analytics intended to inform policy or practice. We envision that such approaches can help education stakeholders identify at-risk students earlier and design more targeted interventions. For instance, a school district could use a similar model on its own data to flag students who might need extra support, while still understanding the context (e.g., low attendance or low sense of belonging) behind each prediction.

In summary, this work contributes to both methodological innovation in the analysis of large-scale assessment data and practical understanding of cross-national patterns in education. We used publicly available PISA 2018 data to benchmark advanced ML models and demonstrated gains in prediction and insight when using explainable AI techniques. The cross-country perspective provided evidence that certain educational challenges and solutions transcend borders (like socio-economic inequities and the need to engage students), while also drawing attention to local context (each country's unique combination of influential factors). We hope these findings will inform researchers, educators, and policymakers in devising strategies that are data-informed yet context-sensitive, ultimately supporting the goal of improving mathematics achievement for all students.

**Declarations**

Supplementary materials related to this work are available in the journal submission system and can be provided by the corresponding author upon reasonable request.

Availability of data and materials: The datasets analyzed in this study are from PISA 2018, which are publicly available through the OECD's website [https://www.oecd.org/en/data/datasets/pisa-2018-database.html]. The processed data and feature engineering code used during the current study are available from the corresponding author on reasonable request. All analysis code (for data preprocessing, model training, and evaluation) is written in Python and can be shared to enable replication of the results.

Competing interests: The authors declare that they have no competing interests. There are no financial or non-financial interests that could be perceived as influencing the research reported.

Funding: This research did not receive any specific grant from funding agencies in the public, commercial, or not-for-profit sectors.

Authors' contributions: Liu conceived the research idea, guided the overall research design, provided expertise on educational theory, interpreted the results in the context of prior literature, contributed to writing the manuscript, and critically revised the content. Rui led the data preprocessing, feature engineering, and machine learning modeling. All authors collaboratively discussed the analytical approach and results. All authors read and approved the final manuscript.

**Table 1**

*Descriptive Sample Characteristics by Country*

| Country | Argentina | Chile | Chinese Taipei | Finland | Hungary | Italy | Japan | Korea | Philippines | US |
|---|---|---|---|---|---|---|---|---|---|---|
| | N (%) | N (%) | N (%) | N (%) | N (%) | N (%) | N (%) | N (%) | N (%) | N (%) |
| Total (raw) N = | 11,975 | 7,621 | 7,243 | 5,649 | 5,132 | 11,785 | 6,109 | 6,650 | 7,233 | 4,838 |
| Gender | 11,975 | 7,621 | 7,243 | 5,649 | 5,132 | 11,785 | 6,109 | 6,650 | 7,233 | 4,838 |
| Female | 6,227 (52%) | 3,814 (50%) | 3,624 (50%) | 2,877 (51%) | 2,605 (51%) | 6,105 (52%) | 3,120 (51%) | 3,459 (52%) | 3,868 (53%) | 2,462 (51%) |
| Male | 5,748 (48%) | 3,807 (50%) | 3,619 (50%) | 2,772 (49%) | 2,527 (49%) | 5,680 (48%) | 2,989 (49%) | 3,191 (48%) | 3,365 (47%) | 2,376 (49%) |
| Use of computers for schoolwork | 11,975 | 7,621 | 7,243 | 5,649 | 5,132 | 11,785 | 6,109 | 6,650 | 7,233 | 4,838 |
| Yes | 8,543 (71%) | 6,338 (83%) | 5,755 (79%) | 5,237 (93%) | 4,683 (91%) | 10,475 (89%) | 3,688 (60%) | 6,003 (90%) | 4,133 (57%) | 4,170 (86%) |
| No | 3,073 (26%) | 1,121 (15%) | 1,401 (19%) | 3,31 (6%) | 384 (7%) | 1,011 (9%) | 2,366 (39%) | 6,20 (9%) | 2,766 (38%) | 6,04 (12%) |
| Number of books at home | 11,975 | 7,621 | 7,243 | 5,649 | 5,132 | 11,785 | 6,109 | 6,650 | 7,233 | 4,838 |
| 0-10 books | 4,451 (37%) | 1,934 (25%) | 1,301 (18%) | 568 (10%) | 619 (12%) | 1,254 (11%) | 838 (14%) | 359 (5%) | 3,782 (52%) | 1,121 (23%) |
| 11-25 books | 2,611 (22%) | 1,871 (25%) | 1,325 (18%) | 740 (13%) | 652 (13%) | 1,893 (16%) | 960 (16%) | 528 (8%) | 1,949 (27%) | 988 (20%) |
| 26-100 books | 2,603 (22%) | 2,114 (28%) | 2,195 (30%) | 1,857 (33%) | 1,256 (24%) | 3,442 (29%) | 2,079 (34%) | 1,750 (26%) | 995 (14%) | 1,380 (29%) |
| 101-200 books | 1,057 (9%) | 866 (11%) | 1,079 (15%) | 1,127 (20%) | 980 (19%) | 2,237 (19%) | 1,072 (18%) | 1,510 (23%) | 235 (3%) | 675 (14%) |
| 201-500 books | 562 (5%) | 448 (6%) | 806 (11%) | 890 (16%) | 856 (17%) | 1,672 (14%) | 756 (12%) | 1,633 (25%) | 97 (1%) | 436 (9%) |
| More than 500 books | 328 (3%) | 194 (3%) | 462 (6%) | 374 (7%) | 721 (14%) | 960 (8%) | 365 (6%) | 847 (13%) | 72 (1%) | 191 (4%) |
| Computer for school work | 11,975 | 7,621 | 7,243 | 5,649 | 5,132 | 11,785 | 6,109 | 6,650 | 7,233 | 4,838 |
| Yes | 8,543 (71%) | 6,338 (83%) | 5,755 (79%) | 5,237 (93%) | 4,683 (91%) | 10,475 (89%) | 3,688 (60%) | 6,003 (90%) | 4,133 (57%) | 4,170 (86%) |
| No | 3,073 (26%) | 1,121 (15%) | 1,401 (19%) | 331 (6%) | 384 (7%) | 1,011 (9%) | 2,366 (39%) | 620 (9%) | 2,766 (38%) | 604 (12%) |
| Mothers Education | 11,975 | 7,621 | 4,838 | 5,649 | 5,132 | 11,785 | 6,109 | 6,650 | 7,233 | 4,838 |
| None | 523 (4%) | 171 (2%) | 111 (2%) | 39 (1%) | 13 (0%) | 58 (0%) | 0 (0%) | 4 (0%) | 409 (6%) | 74 (2%) |
| Primary education | 1,453 (12%) | 185 (2%) | 156 (2%) | 17 (0%) | 18 (0%) | 151 (1%) | 0 (0%) | 57 (1%) | 689 (10%) | 104 (2%) |
| Lower secondary education | 2,088 (17%) | 946 (12%) | 579 (8%) | 158 (3%) | 433 (8%) | 2491 (21%) | 155 (3%) | 300 (5%) | 854 (12%) | 387 (8%) |
| Vocational/pre-vocational upper secondary | 0 (0%) | 712 (9%) | 0 (0%) | 0 (0%) | 743 (14%) | 602 (5%) | 257 (4%) | 492 (7%) | 457 (6%) | 0 (0%) |
| Upper secondary or non-tertiary post-secondary | 2,116 (18%) | 1,980 (26%) | 2,166 (30%) | 1,243 (22%) | 1,482 (29%) | 4,203 (36%) | 2,215 (36%) | 1,821 (27%) | 2,161 (30%) | 1,447 (30%) |
| Vocational tertiary | 1,865 (16%) | 868 (11%) | 1,030 (14%) | 1,062 (19%) | 273 (5%) | 684 (6%) | 1,654 (27%) | 332 (5%) | 986 (14%) | 688 (14%) |
| Tertiary education | 3,463 (29%) | 2,382 (31%) | 3,062 (42%) | 2,980 (53%) | 2,099 (41%) | 3,195 (27%) | 1,694 (28%) | 3,605 (54%) | 1,605 (22%) | 2,045 (42%) |
| Fathers Education | 11,975 | 7,621 | 7,243 | 5,649 | 5,132 | 11,785 | 6,109 | 6,650 | 7,233 | 4,838 |
| None | 740 (6%) | 177 (2%) | 39 (1%) | 63 (1%) | 7 (0%) | 75 (1%) | 0 (0%) | 11 (0%) | 647 (9%) | 93 (2%) |
| Primary education | 1,450 (12%) | 179 (2%) | 228 (3%) | 48 (1%) | 15 (0%) | 250 (2%) | 0 (0%) | 45 (1%) | 693 (10%) | 118 (2%) |
| Lower secondary education | 2,353 (20%) | 1,017 (13%) | 1,081 (15%) | 269 (5%) | 426 (8%) | 2934 (25%) | 292 (5%) | 189 (3%) | 707 (10%) | 448 (9%) |
| Vocational/pre-vocational upper secondary | 0 (0%) | 779 (10%) | 969 (13%) | 0 (0%) | 930 (18%) | 557 (5%) | 385 (6%) | 526 (8%) | 436 (6%) | 0 (0%) |
| Upper secondary or non-tertiary post-secondary | 2,191 (18%) | 1,563 (21%) | 1,276 (18%) | 1,424 (25%) | 1,435 (28%) | 4,124 (35%) | 2,021 (33%) | 1,585 (24%) | 2,234 (31%) | 1,784 (37%) |
| Vocational tertiary | 1,441 (12%) | 732 (10%) | 1,964 (27%) | 1,160 (21%) | 312 (6%) | 625 (5%) | 657 (11%) | 286 (4%) | 865 (12%) | 535 (11%) |
| Tertiary education | 2,783 (23%) | 2,558 (34%) | 1,553 (21%) | 2,423 (43%) | 1,885 (37%) | 2,696 (23%) | 2,389 (39%) | 3,933 (59%) | 1,519 (21%) | 1,674 (35%) |
| Highest parental education | 11,975 | 7,621 | 7,243 | 5,649 | 5,132 | 11,785 | 6,109 | 6,650 | 7,233 | 4,838 |
| None | 210 (2%) | 46 (1%) | 10 (0%) | 15 (0%) | 3 (0%) | 18 (0%) | 0 (0%) | 0 (0%) | 217 (3%) | 39 (1%) |
| Primary education | 1,098 (9%) | 92 (1%) | 81 (1%) | 10 (0%) | 5 (0%) | 46 (0%) | 0 (0%) | 9 (0%) | 428 (6%) | 70 (1%) |
| Lower secondary education | 1,753 (15%) | 698 (9%) | 552 (8%) | 69 (1%) | 256 (5%) | 1653 (14%) | 83 (1%) | 83 (1%) | 664 (9%) | 276 (6%) |
| Vocational/pre-vocational upper secondary | 0 (0%) | 619 (8%) | 835 (12%) | 0 (0%) | 629 (12%) | 488 (4%) | 143 (2%) | 238 (4%) | 354 (5%) | 0 (0%) |
| Upper secondary or non-tertiary post-secondary | 2,341 (20%) | 1,769 (23%) | 1,492 (21%) | 998 (18%) | 1,387 (27%) | 4,310 (37%) | 1,762 (29%) | 1,523 (23%) | 2,405 (33%) | 1,274 (26%) |
| Vocational tertiary | 1,806 (15%) | 920 (12%) | 2,329 (32%) | 929 (16%) | 280 (5%) | 790 (7%) | 1,182 (19%) | 257 (4%) | 1,112 (15%) | 647 (13%) |
| Tertiary education | 4,450 (37%) | 3,200 (42%) | 1,855 (26%) | 3,503 (62%) | 2,516 (49%) | 4,134 (35%) | 2,845 (47%) | 4,513 (68%) | 2,004 (28%) | 2,453 (51%) |
| Parental support | 11,975 | 7,621 | 7,243 | 5,649 | 5,132 | 11,785 | 6,109 | 6,650 | 7,233 | 4,838 |
| Strongly disagree | 3,951 (33%) | 288 (4%) | 188 (3%) | 158 (3%) | 181 (4%) | 699 (6%) | 301 (5%) | 61 (1%) | 416 (6%) | 152 (3%) |
| Disagree | 592 (5%) | 287 (4%) | 424 (6%) | 332 (6%) | 292 (6%) | 946 (8%) | 608 (10%) | 225 (3%) | 316 (4%) | 192 (4%) |
| Agree | 3,339 (28%) | 1,817 (24%) | 3,886 (54%) | 2,285 (40%) | 1,893 (37%) | 3,934 (33%) | 2,763 (45%) | 2,952 (44%) | 2,781 (38%) | 1,603 (33%) |
| Strongly agree | 3,951 (33%) | 3,005 (39%) | 2,584 (36%) | 2,392 (42%) | 2,085 (41%) | 3,691 (31%) | 2,307 (38%) | 3,353 (50%) | 3,095 (43%) | 2,674 (55%) |
| Teacher's clear goal setting | 11,975 | 7,621 | 7,243 | 5,649 | 5,132 | 11,785 | 6,109 | 6,650 | 7,233 | 4,838 |
| Every lesson | 5,047 (42%) | 4,375 (57%) | 2,180 (30%) | 1,671 (30%) | 1,618 (32%) | 3,725 (32%) | 2,020 (33%) | 2,837 (43%) | 3,990 (55%) | 1,929 (40%) |
| Most lessons | 3,689 (31%) | 2,121 (28%) | 2,607 (36%) | 2,475 (44%) | 1,110 (22%) | 4,281 (36%) | 2,519 (41%) | 2,592 (39%) | 1,974 (27%) | 1,693 (35%) |
| Some lessons | 2,135 (18%) | 713 (9%) | 1,937 (27%) | 1,052 (19%) | 1,919 (37%) | 2,328 (20%) | 1,086 (18%) | 986 (15%) | 953 (13%) | 868 (18%) |
| Never or hardly ever | 657 (5%) | 158 (2%) | 435 (6%) | 277 (5%) | 408 (8%) | 1026 (9%) | 431 (7%) | 198 (3%) | 199 (3%) | 228 (5%) |
| Feedback on student strengths | 11,975 | 7,621 | 7,243 | 5,649 | 5,132 | 11,785 | 6,109 | 6,650 | 7,233 | 4,838 |
| Never or almost never | 3,830 (32%) | 1,761 (23%) | 1,156 (16%) | 1,258 (22%) | 1,394 (27%) | 4,703 (40%) | 3,149 (52%) | 1,740 (26%) | 707 (10%) | 779 (16%) |
| Some lessons | 4,813 (40%) | 2,490 (33%) | 2,701 (37%) | 2,457 (43%) | 2,026 (39%) | 3,922 (33%) | 1,742 (29%) | 1,971 (30%) | 3,339 (46%) | 1,759 (36%) |
| Many lessons | 1,944 (16%) | 1,967 (26%) | 2,215 (31%) | 1,365 (24%) | 1,247 (24%) | 2,020 (17%) | 798 (13%) | 1,812 (27%) | 1,852 (26%) | 1,345 (28%) |
| Every lesson or almost every lesson | 804 (7%) | 1072 (14%) | 1081 (15%) | 383 (7%) | 379 (7%) | 694 (6%) | 347 (6%) | 1085 (16%) | 1222 (17%) | 827 (17%) |
| Teacher adaptation to class needs | 11,975 | 7,621 | 7,243 | 5,649 | 5,132 | 11,785 | 6,109 | 6,650 | 7,233 | 4,838 |
| Never or almost never | 1,744 (15%) | 619 (8%) | 739 (10%) | 579 (10%) | 578 (11%) | 1506 (13%) | 612 (10%) | 321 (5%) | 509 (7%) | 484 (10%) |
| Some lessons | 4,579 (38%) | 1,931 (25%) | 2,713 (37%) | 2,133 (38%) | 1,575 (31%) | 3,873 (33%) | 1,223 (20%) | 1,321 (20%) | 2,757 (38%) | 1,944 (40%) |
| Many lessons | 3,281 (27%) | 2,708 (36%) | 2,580 (36%) | 1,967 (35%) | 1,815 (35%) | 3,774 (32%) | 2,886 (47%) | 2,967 (45%) | 2,101 (29%) | 1,460 (30%) |
| Every lesson or almost every lesson | 1,869 (16%) | 2,083 (27%) | 1,131 (16%) | 783 (14%) | 1,094 (21%) | 2,191 (19%) | 1,340 (22%) | 1,998 (30%) | 1,766 (24%) | 819 (17%) |
| Competitive spirit among students | 11,975 | 7,621 | 7,243 | 5,649 | 5,132 | 11,785 | 6,109 | 6,650 | 7,233 | 4,838 |
| Not at all true | 1,511 (13%) | 550 (7%) | 338 (5%) | 316 (6%) | 579 (11%) | 1,479 (13%) | 987 (16%) | 458 (7%) | 378 (5%) | 272 (6%) |
| Slightly true | 3,746 (31%) | 2,128 (28%) | 2,124 (29%) | 1,796 (32%) | 2,060 (40%) | 4,231 (36%) | 2,918 (48%) | 1,336 (20%) | 2,463 (34%) | 1,399 (29%) |
| Very true | 2,173 (18%) | 1,777 (23%) | 3,182 (44%) | 2,448 (43%) | 1,361 (27%) | 2,542 (22%) | 1,563 (26%) | 2,658 (40%) | 2,782 (38%) | 1,875 (39%) |
| Extremely true | 835 (7%) | 833 (11%) | 1431 (20%) | 540 (10%) | 388 (8%) | 795 (7%) | 490 (8%) | 2131 (32%) | 903 (12%) | 1,027 (21%) |
| Cooperative attitudes among students | 11,975 | 7,621 | 7,243 | 5,649 | 5,132 | 11,785 | 6,109 | 6,650 | 7,233 | 4,838 |
| Not at all true | 932 (8%) | 295 (4%) | 296 (4%) | 197 (3%) | 288 (6%) | 750 (6%) | 310 (5%) | 469 (7%) | 244 (3%) | 262 (5%) |
| Slightly true | 3,866 (32%) | 2,015 (26%) | 1,731 (24%) | 1,288 (23%) | 1,592 (31%) | 3,545 (30%) | 1,818 (30%) | 1,613 (24%) | 1,902 (26%) | 1,833 (38%) |
| Very true | 2,385 (20%) | 1,943 (26%) | 3,631 (50%) | 2,944 (52%) | 1,876 (37%) | 3,501 (30%) | 2,643 (43%) | 3,405 (51%) | 3,079 (43%) | 2,037 (42%) |
| Extremely true | 695 (6%) | 671 (9%) | 1,407 (19%) | 519 (9%) | 494 (10%) | 775 (7%) | 1,111 (18%) | 1,101 (17%) | 1,046 (14%) | 404 (8%) |
| Personal learning goals | 11,975 | 7,621 | 7,243 | 5,649 | 5,132 | 11,785 | 6,109 | 6,650 | 7,233 | 4,838 |
| Not at all true of me | 661 (6%) | 166 (2%) | 327 (5%) | 159 (3%) | 694 (14%) | 565 (5%) | 400 (7%) | 463 (7%) | 269 (4%) | 104 (2%) |
| Slightly true of me | 2,630 (22%) | 455 (6%) | 1,486 (21%) | 493 (9%) | 947 (18%) | 1,611 (14%) | 1,400 (23%) | 1,516 (23%) | 839 (12%) | 384 (8%) |
| Moderately true of me | 2,553 (21%) | 1,705 (22%) | 2,891 (40%) | 1,667 (30%) | 2,211 (43%) | 3,711 (31%) | 2,399 (39%) | 1,791 (27%) | 1,050 (15%) | 1,384 (29%) |
| Very true of me | 2,207 (18%) | 2,076 (27%) | 1,430 (20%) | 2,146 (38%) | 765 (15%) | 3,096 (26%) | 1,186 (19%) | 1,748 (26%) | 2,756 (38%) | 1,534 (32%) |
| Extremely true of me | 2,499 (21%) | 2,204 (29%) | 999 (14%) | 841 (15%) | 327 (6%) | 1,750 (15%) | 635 (10%) | 1,079 (16%) | 2,106 (29%) | 1,238 (26%) |
| Perceived teacher motivation | 11,975 | 7,621 | 7,243 | 5,649 | 5,132 | 11,785 | 6,109 | 6,650 | 7,233 | 4,838 |
| Never or hardly ever | 1,744 (15%) | 627 (8%) | 598 (8%) | 558 (10%) | 503 (10%) | 1,597 (14%) | 532 (9%) | 467 (7%) | 317 (4%) | 440 (9%) |
| In some lessons | 3,616 (30%) | 1,894 (25%) | 2,725 (38%) | 1,872 (33%) | 1,466 (29%) | 3,604 (31%) | 1,507 (25%) | 1,690 (25%) | 1,895 (26%) | 1,455 (30%) |
| In most lessons | 3,545 (30%) | 2,563 (34%) | 2,525 (35%) | 2,078 (37%) | 1,987 (39%) | 3,997 (34%) | 2,164 (35%) | 2,780 (42%) | 2,366 (33%) | 1,662 (34%) |
| In all lessons | 2,521 (21%) | 2,162 (28%) | 1,300 (18%) | 907 (16%) | 1,063 (21%) | 2,034 (17%) | 1,840 (30%) | 1,663 (25%) | 2,487 (34%) | 1,135 (23%) |

**Table 2**

*Predictive Performance of Four Models Across Ten Countries ($R^2$ / MAE)*

| Country | MLR | Random Forest | CATBoost | ANN |
|---|---|---|---|---|
| Argentina | 0.26 / 56.2 | 0.29 / 54.9 | 0.27 / 55.7 | 0.26 / 56.4 |
| Chile | 0.28 / 58.1 | 0.31 / 56.4 | 0.29 / 57.4 | 0.23 / 59.0 |
| Chinese Taipei | 0.24 / 69.2 | 0.31 / 66.8 | 0.26 / 68.6 | 0.17 / 72.5 |
| Finland | 0.17 / 57.2 | 0.21 / 55.6 | 0.19 / 56.4 | 0.15 / 59.2 |
| Hungary | 0.29 / 57.3 | 0.38 / 53.9 | 0.33 / 56.3 | 0.31 / 57.5 |
| Italy | 0.19 / 62.1 | 0.25 / 59.8 | 0.22 / 61.1 | 0.16 / 62.6 |
| Japan | 0.18 / 63.7 | 0.32 / 57.4 | 0.28 / 59.3 | 0.26 / 64.3 |
| Korea | 0.17 / 70.0 | 0.24 / 67.0 | 0.22 / 68.1 | 0.17 / 69.9 |
| Philippines | 0.28 / 50.4 | 0.38 / 46.6 | 0.36 / 47.5 | 0.29 / 49.2 |
| United States | 0.20 / 64.7 | 0.24 / 63.5 | 0.22 / 64.2 | 0.27 / 62.7 |

*Note.* $R^2$ = proportion of variance in maths scores explained by the model (higher is better). MAE = Mean Absolute Error in points (lower is better). Values represent the coefficient of determination ($R^2$) and MAE on the test set for each country. RF and CATBoost models generally outperform MLR and ANN in both accuracy and error across contexts. All metrics are computed on each country's test set.

**Figure 1**

*Distributions of PISA 2018 Mathematics Scores in The Ten Study Countries*

Argentina

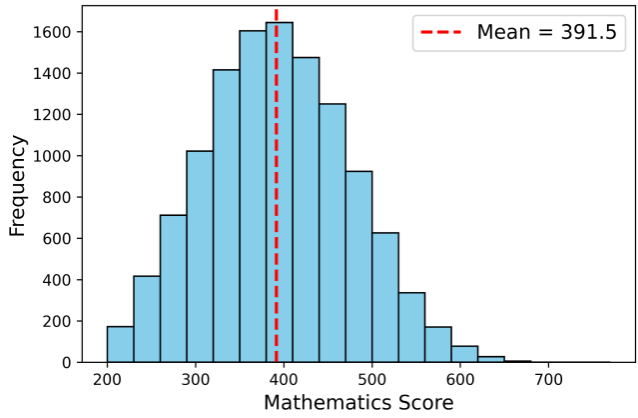

Chile

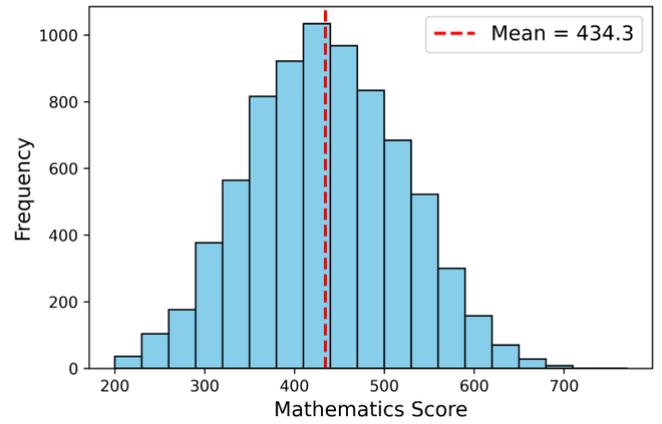

Chinese Taipei

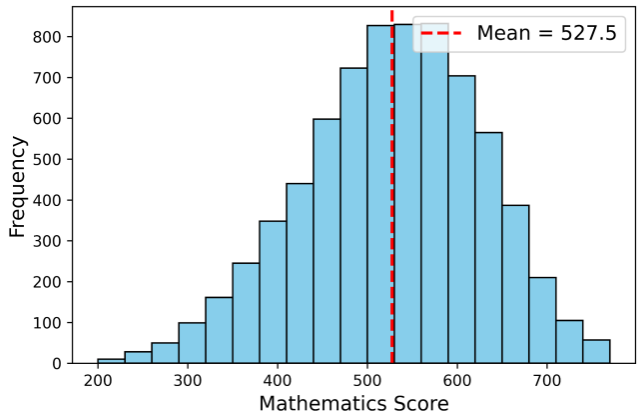

Finland

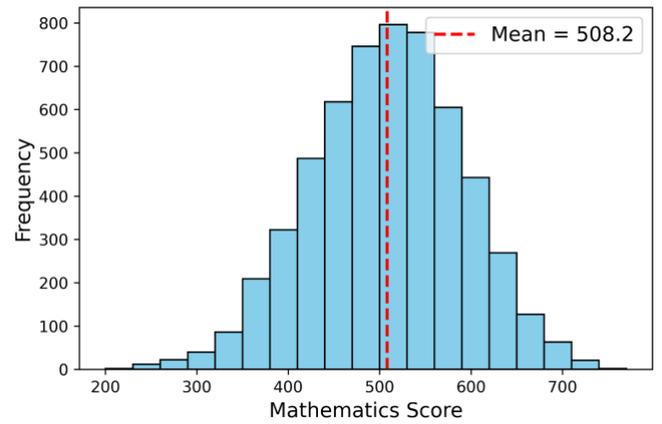

Hungary

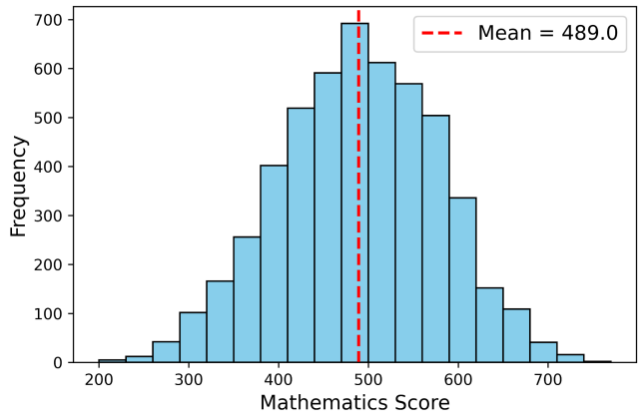

Italy

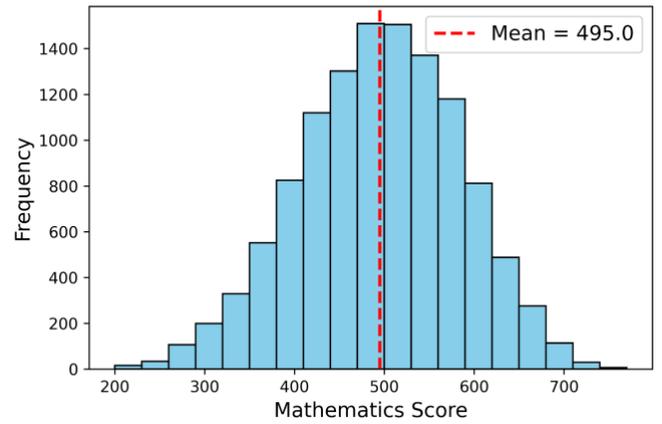

Japan

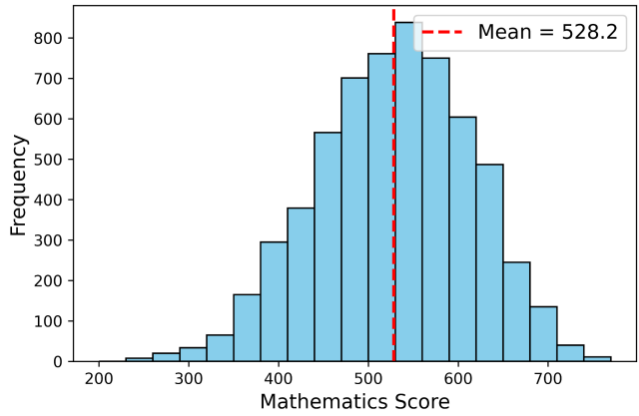

Korea

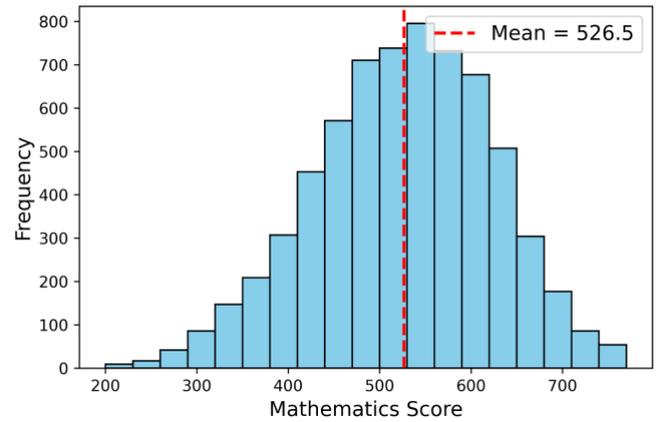

Philippines

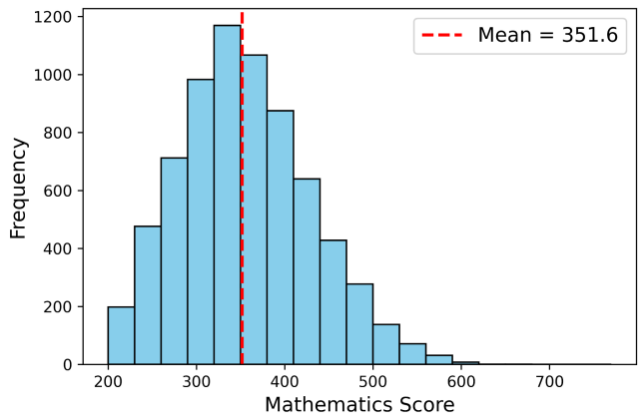

U.S.

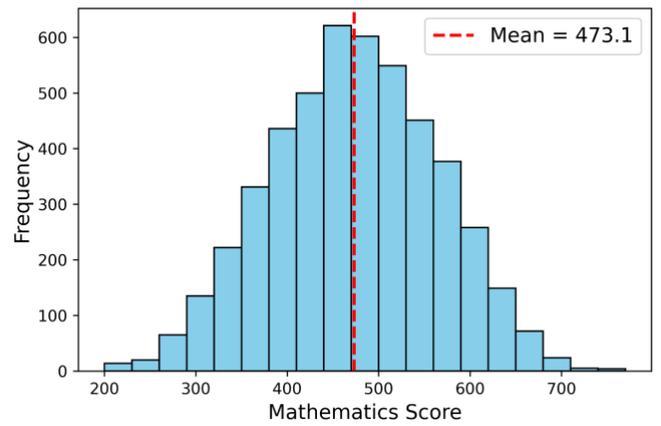

*Note.* Each country's distribution is shown with its mean indicated by a vertical line. These distributions illustrate differences in central tendency and spread, highlighting contexts with higher overall performance (e.g. East Asia) versus those with greater proportions of low-achieving students.

**Figure 2**

*Top Predictors of Mathematics Performance, Based on Boosting Feature Importance*

Argentina

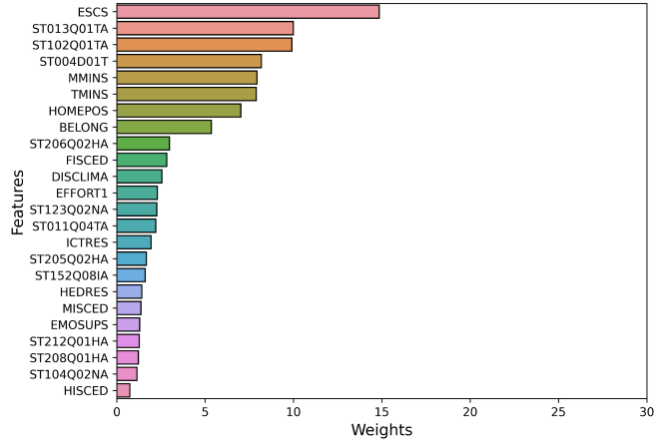

Chile

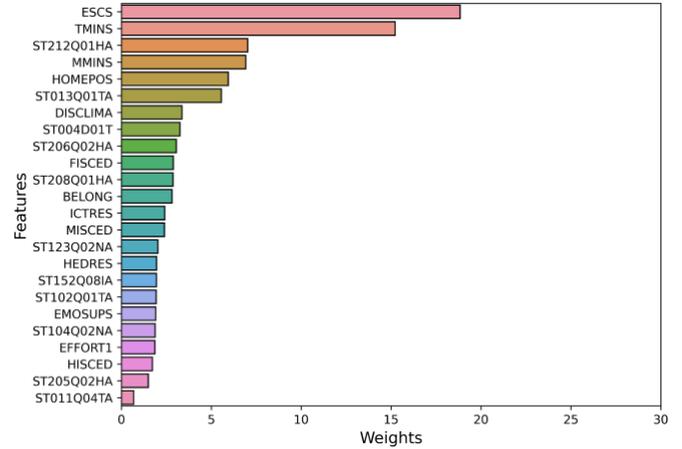

Chinese Taipei

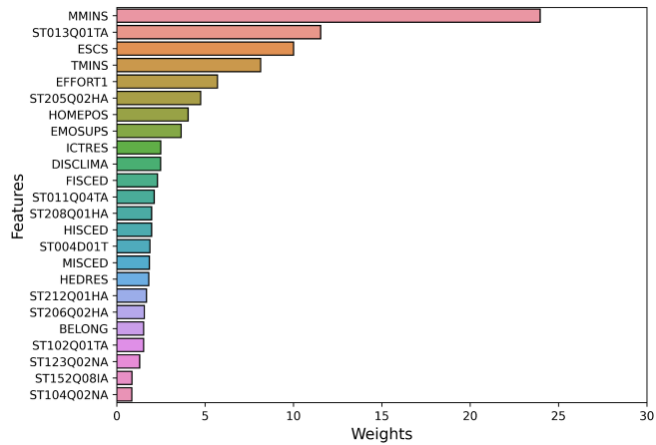

Finland

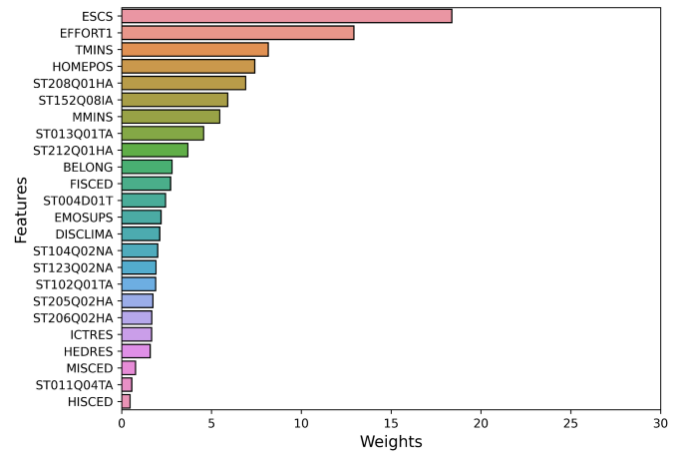

## Hungary

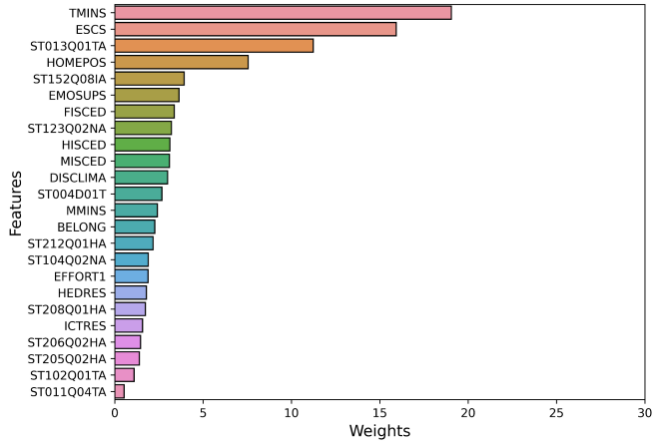

## Italy

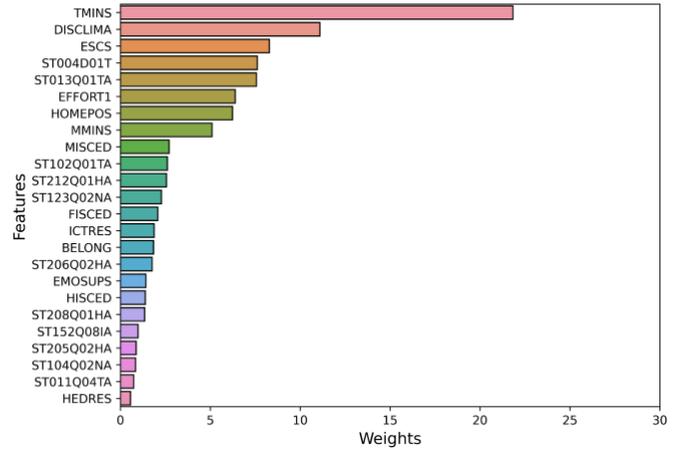

## Japan

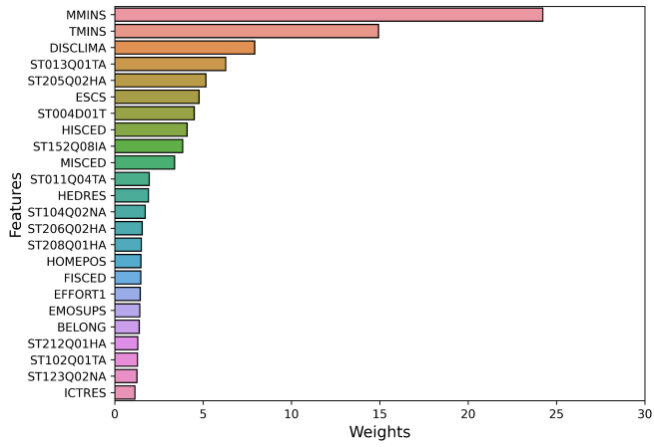

## Korea

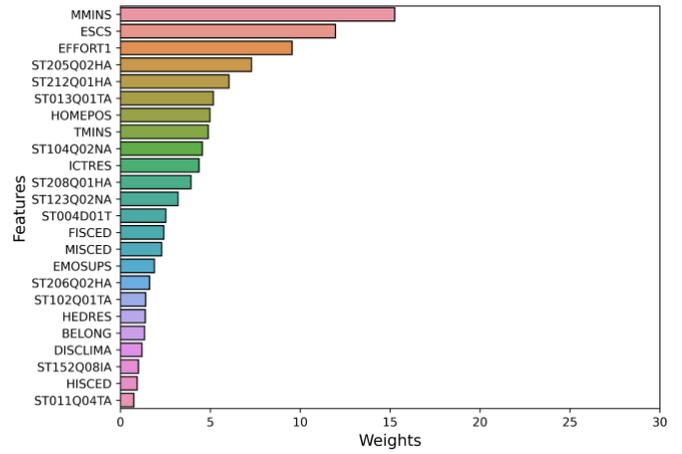

Philippines

U.S.

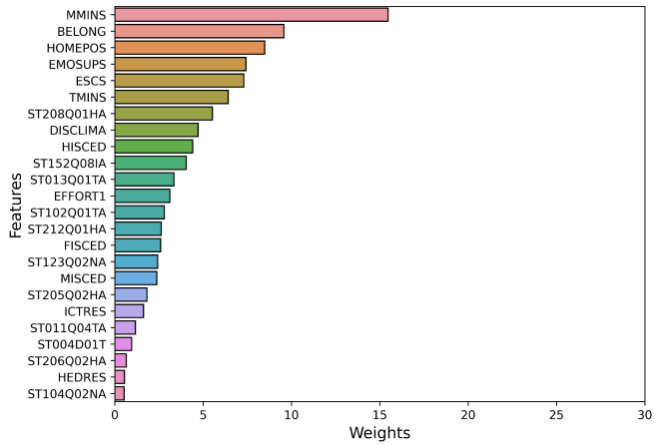
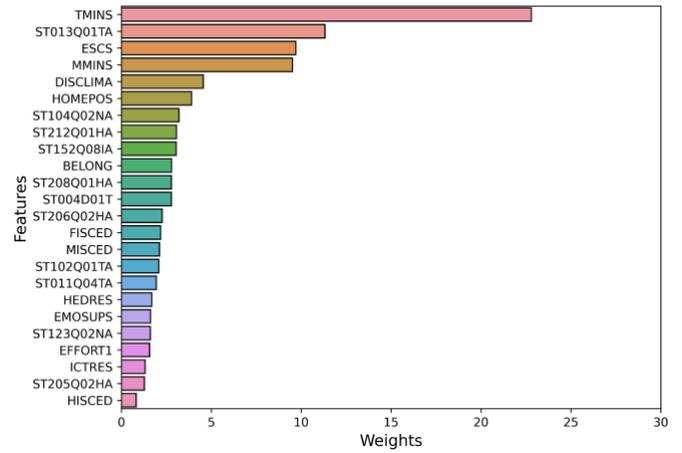

*Note.* This figure illustrates the top predictors of mathematics achievement across all countries, based on a combined SHAP value analysis. Features are grouped by category (student, family, school) and ordered by overall importance. Colored bars indicate the average magnitude of influence on the predicted score. Socio-economic status (ESCS) is the top feature, followed by student effort and various engagement indices, while individual demographic factors like gender have relatively smaller effects.

**Figure 3**

*Distribution of the Three Most Important Predictive Features by Variable Level and Country.*

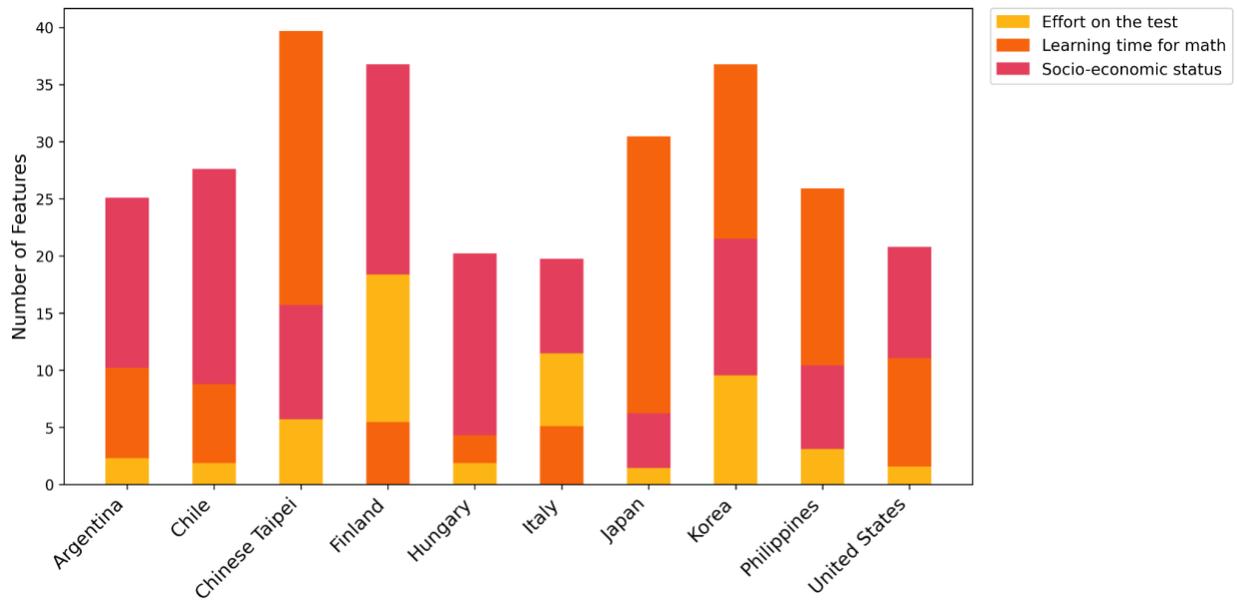

*Note.* This bar chart shows how the top-ranked predictors (based on SHAP values) differ across countries. Socio-economic status, student effort on the test, and learning time in mathematics dominate across contexts but in varying proportions, highlighting both shared and country-specific drivers of achievement.

**Figure 4.** *Distribution of All Predictive Features by Variable Level and Country*

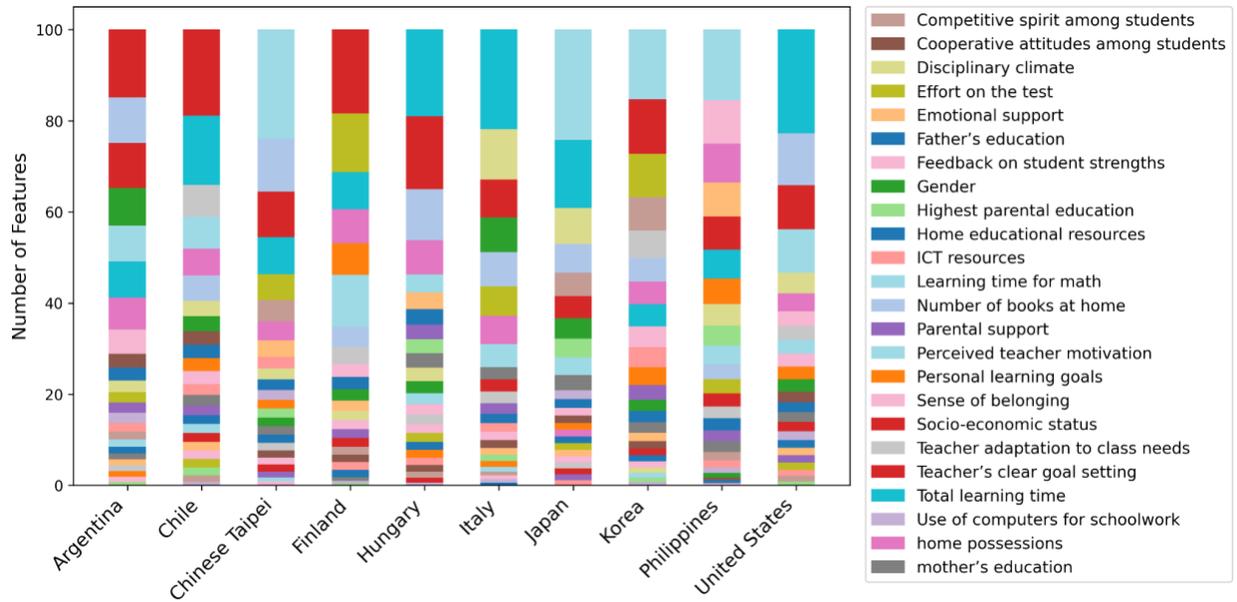

*Note.* This stacked bar chart displays the number of times each predictor was selected as important across countries, grouped by variable name. It shows broad diversity in feature importance profiles and emphasizes which student, family, and school factors are most consistently relevant.

**Figure 5**

*Decision Tree Visualization of Predictors Across Ten Countries*

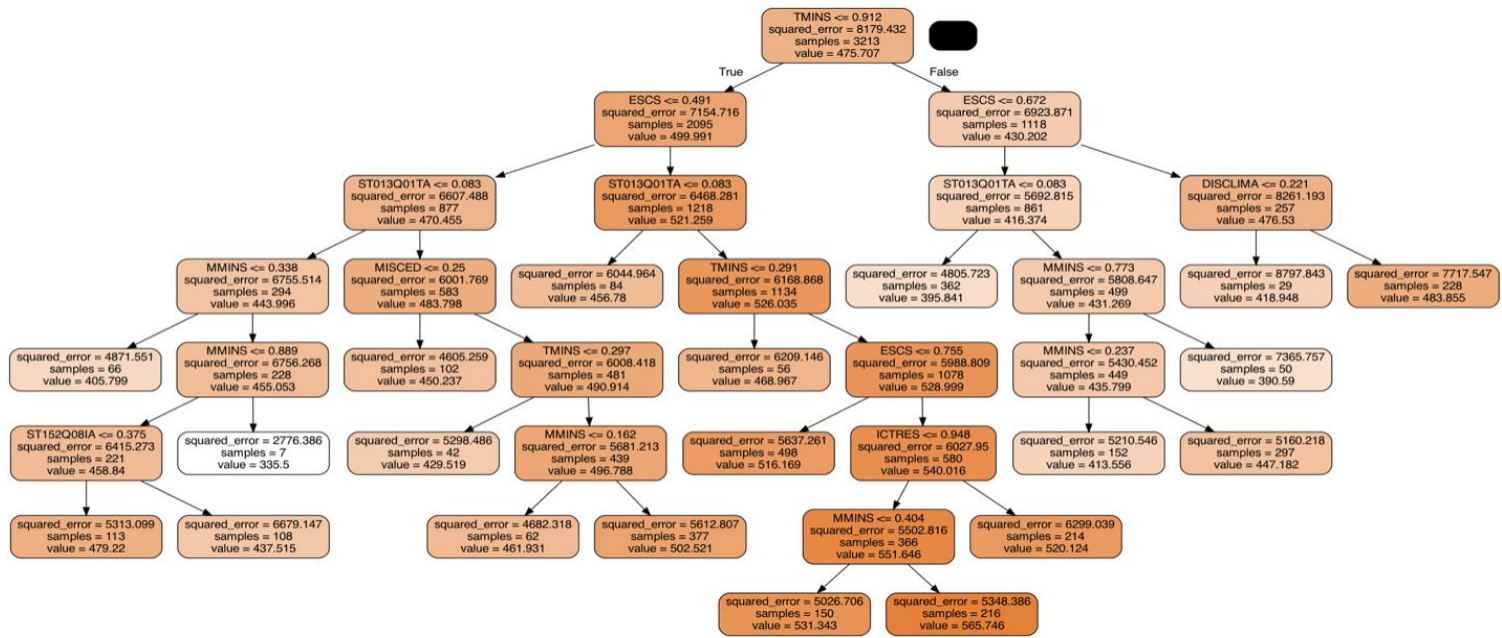

*Note.* Decision tree example extracted from the Random Forest model. This figure provides an interpretable snapshot of how the model makes predictions. Each node shows a splitting rule (e.g. "TMINS < 10 hours/week?") and each branch leads to a predicted score range. The tree highlights interactions, for instance: students with very low study time split off early with lower scores, and within that group, SES further stratifies outcomes, while in higher study time branches, factors like teacher motivation appear. This visualization helps illustrate the conditional logic the model has learned.

**Figure 6**

*Hexbin Plots of Predicted vs. Actual Scores for ANN And Catboost (Argentina Example)*

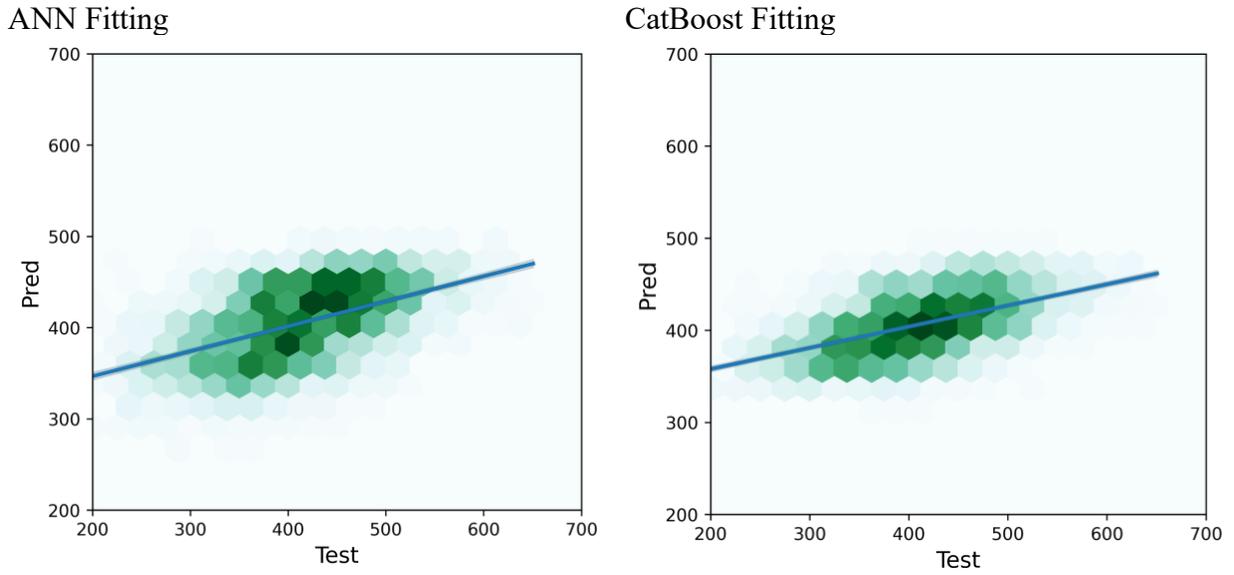

*Note.* Each hexagon's color density represents the concentration of students at a given combination of actual and predicted score. A perfect prediction would lie along the diagonal line (actual = predicted).

**Figure 7**

*Distribution of Predicted Mathematics Scores in Training and Test Sets by Country (Random Forest Model)*

Argentina

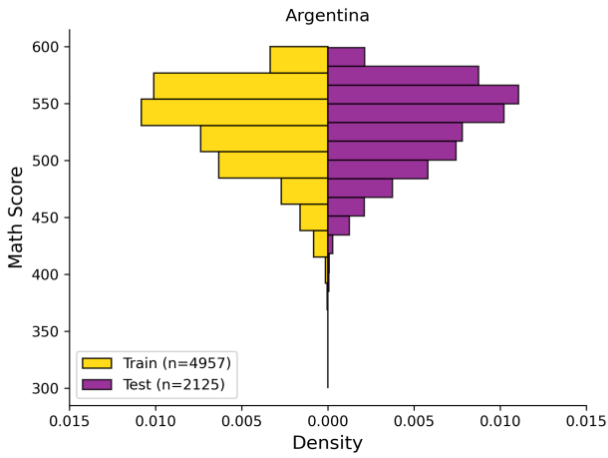

Chile

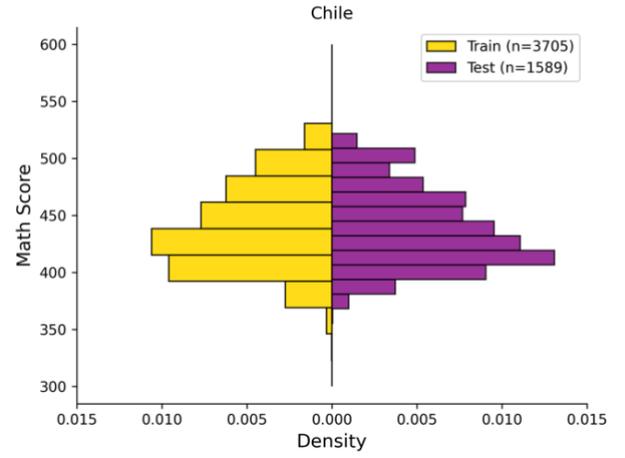

Chinese Taipei

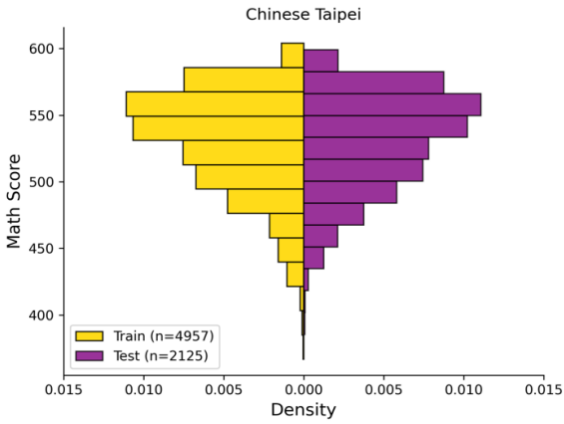

Finland

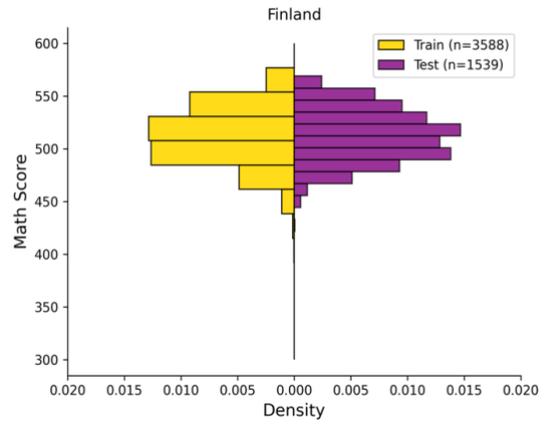

Hungary

Italy

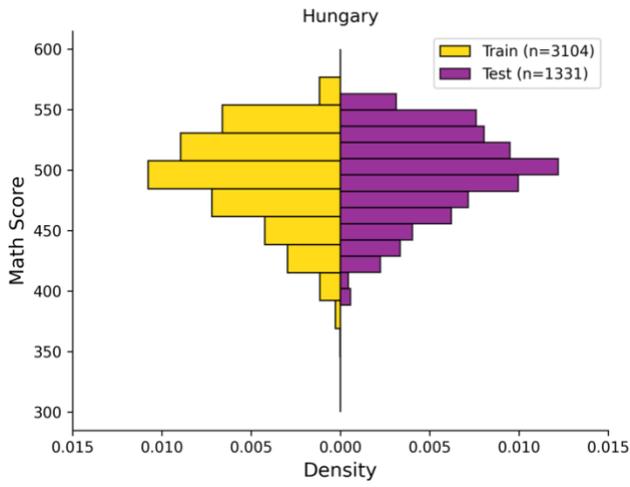

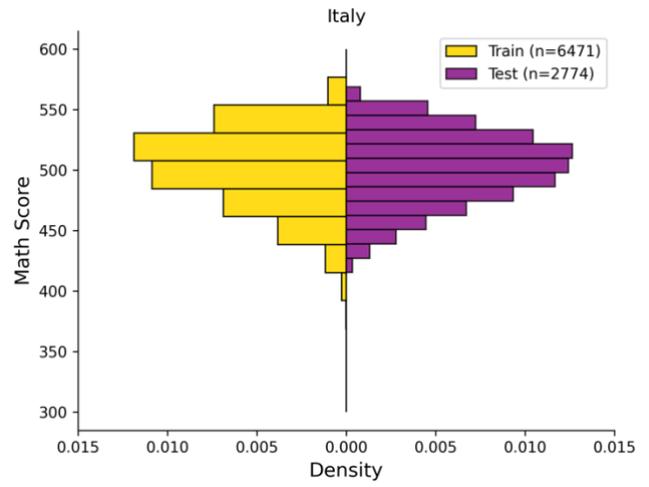

Japan

Korea

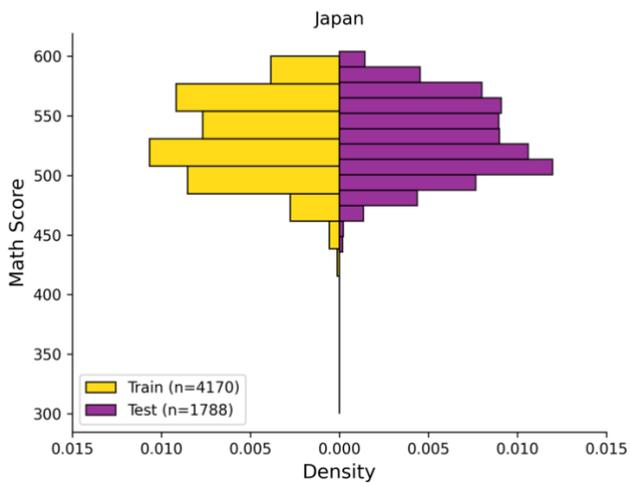

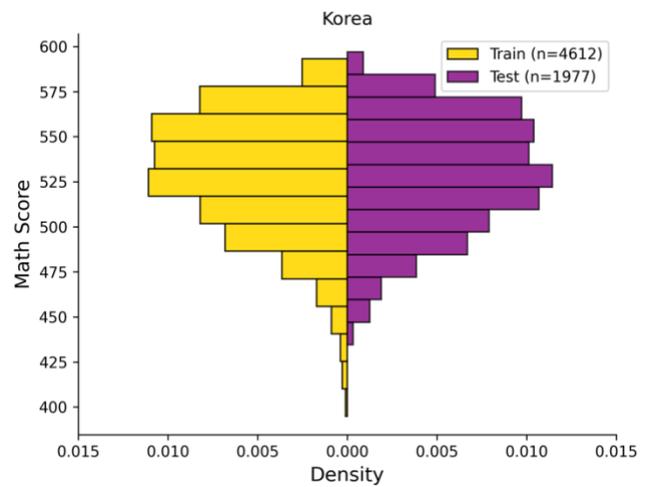

Philippines                                        U.S.

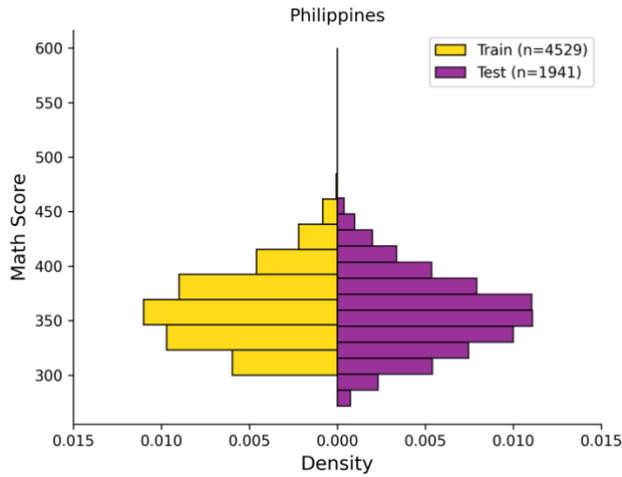 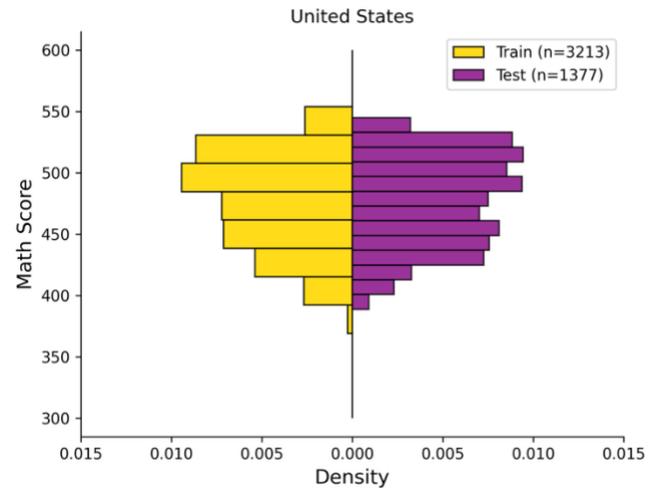

*Note.* Yellow bars represent the predicted scores from the training set; purple bars show predicted scores from the test set. The close alignment between distributions suggests that the Random Forest model generalizes well, with minimal overfitting. Slight divergence (e.g., broader or shifted test distributions) reflects context-specific limits to model generalization.